\documentclass[twoside,11pt]{article}
\usepackage{comment}
\usepackage{blindtext}
\usepackage{wrapfig}
\usepackage{amsmath, amsthm, amssymb, amsfonts}
\usepackage{graphicx}
\usepackage{algorithm}
\usepackage{algorithmic}

\usepackage{booktabs}      
\usepackage{amsfonts}      
\usepackage{nicefrac}      

\newtheorem{Example}{Example}
\newtheorem{Theorem}{Theorem}
\newtheorem{Lemma}{Lemma}
\newtheorem{Definition}{Definition}
\newtheorem{Proposition}{Proposition}

\usepackage{jmlr2e}

\usepackage{lastpage}
\jmlrheading{23}{2024}{1-\pageref{LastPage}}{month/day; Revised x/x}{x/x}{21-0000}{Tianyu Ruan and Shihua Zhang}

\ShortHeadings{Towards Understanding how attention mechanism works in deep learning}{Tianyu Ruan and Shihua Zhang}
\firstpageno{1}

\begin{document}
\title{Towards understanding how attention mechanism works in deep learning}

\author{\name Tianyu Ruan$^{\star\,*}$ \email ruantianyu17@mails.ucas.ac.cn 
       \AND
       \name Shihua Zhang$^{\star\,*}$ \email zsh@amss.ac.cn \\
       \addr$^{\star}$ Academy of Mathematics and Systems Science\\
       Chinese Academy of Sciences\\
       $^*$ School of Mathematical Sciences\\
       University of Chinese Academy of Sciences}

\editor{My editor}

\maketitle

\begin{abstract}
Attention mechanism has been extensively integrated within mainstream neural network architectures, such as Transformers and graph attention networks. Yet, its underlying working principles remain somewhat elusive. What is its essence? Are there any connections between it and traditional machine learning algorithms? In this study, we inspect the process of computing similarity using classic metrics and vector space properties in manifold learning, clustering, and supervised learning. We identify the key characteristics of similarity computation and information propagation in these methods and demonstrate that the self-attention mechanism in deep learning adheres to the same principles but operates more flexibly and adaptively. We decompose the self-attention mechanism into a learnable pseudo-metric function and an information propagation process based on similarity computation. We prove that the self-attention mechanism converges to a drift-diffusion process through continuous modeling provided the pseudo-metric is a transformation of a metric and certain reasonable assumptions hold. This equation could be transformed into a heat equation under a new metric. In addition, we give a first-order analysis of attention mechanism with a general pseudo-metric function. This study aids in understanding the effects and principle of attention mechanism through physical intuition. Finally, we propose a modified attention mechanism called metric-attention by leveraging the concept of metric learning to facilitate the ability to learn desired metrics more effectively. Experimental results demonstrate that it outperforms self-attention regarding training efficiency, accuracy, and robustness. 
\end{abstract}
 
\subsection*{Keywords}
Attention mechanism, Transformer, graph attention network, similarity computation, heat diffusion

\section{Introduction}
The attention or self-attention mechanism is extensively applied in popular deep learning architectures like Transformers \citep{vaswani2017attention,dong2018speech} and graph attention networks \citep{velivckovic2017graph}. This mechanism enables the model to assign diverse weights to various parts (data points, nodes, etc.) of the input sequence (data, graph, etc.) based on their relevance when producing an output. This capability is particularly critical for handling inputs where the lengths and relevance strengths of different parts can vary significantly. As a result, this mechanism contributes to the broad applications of deep learning in various fields, including natural language processing \citep{devlin2018bert,brown2020language,radford2019language}, computer vision  \citep{dosovitskiy2020image,touvron2021training,carion2020end}, graph mining \citep{liu2023cross} and bioinformatics \citep{dong2022deciphering,zhang2023applications,ji2021dnabert}.
 
However, understanding the mathematical principle of attention mechanism is still challenging due to its interaction with normalization layers and feed-forward networks in neural network architectures. Difficulties in understanding attention mechanism also stem from the numerous parameters in neural networks and various engineering techniques. To our knowledge, only a few studies have explored it in depth \citep{vuckovic2020mathematical,dong2021attention,sander2022sinkformers,geshkovski2023mathematical}.  \cite{sander2022sinkformers} formalized the self-attention mechanism with residual connections as a flow map and analyzed it from the perspective of the Wasserstein gradient flow. In addition, they characterized the L2 self-attention \citep{kim2021lipschitz} using continuous dynamical systems. \cite{geshkovski2023mathematical} investigated attention mechanism in Transformers by assuming that data is distributed on the unit sphere and making simplified assumptions about parameters and proved that the distribution would converge to a single point under certain conditions, suggesting that attention mechanism induces an aggregation tendency. However, they did not fully explain how attention mechanism works or its connection to classical algorithms.

Moreover, several architectures, such as CRATE \citep{yu2024white} and Probabilistic Transformer \citep{Wu_2023}, have been proposed. They often originate from interpretable models and have information propagation mechanisms similar to attention mechanism. While not strictly equivalent, they offer valuable insights into understanding attention mechanism. For example, CRATE suggests that attention mechanism functions as a compression process, whereas the Probabilistic Transformer explains it as an explicit iteration of the Frank-Wolfe optimization algorithm.

\begin{wrapfigure}{r}{0.4\textwidth}
   \center
   \includegraphics[scale=0.9]{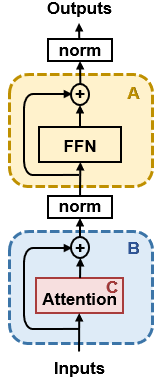}
   \caption{Illustration of three architecture components including the residual block (A), the attention block (C), and their recombination (B) of a Transformer block.}
   \label{fig:our_work}
\end{wrapfigure}

We illustrate three architecture components, i.e.,  the residual block, the attention block, and their recombination of a Transformer block (Figure \ref{fig:our_work}). The attention block consists of an information propagation process followed by a linear transformation. Some studies modeled the residual block using an ordinary differential equation \citep{articleE,chen2018neural}. Some researchers modeled the recombination of the residual and attention blocks using the flow map mentioned above. In this paper, we focus on the attention-based information propagation mechanism. This mechanism can be viewed as a message-passing process akin to that in graph neural networks (GNNs) operating on fully connected graphs. However, GNNs typically employ fixed edge weights and topologies, which limit theoretical analysis to diffusion processes on graphs \citep{li2024generalized}. In contrast, attention mechanism, as a learnable method of information propagation on fully connected graphs, has yet to be thoroughly analyzed in terms of its limit behavior.

Many machine learning methods such as manifold learning (e.g., diffusion map \citep{coifman2006diffusion}, UMAP \citep{mcinnes2018umap}), clustering methods (e.g., k-means clustering \citep{lloyd1982least}, fuzzy c-means clustering \citep{bezdek1984fcm}, Markov clustering \citep{van2008graph}) and supervised learning (e.g., k-nearest neighbors algorithm \citep{fix1985discriminatory}, support vector machine \citep{cortes1995support}),
involve computing similarity or dissimilarity for pairwise data points (or nodes in a graph). These approaches usually consist of some of the following components successively: (1) Initializing similarity based on the input data and the adopted pseudo-metric; (2) Strengthening similarity through some transformation to make similar data points more similar or dissimilar ones more distinct; (3) Normalizing similarity to transform the similarity matrix into a type of probability distribution, which allows the use of probabilistic tools for comparison or further manipulation. Traditional algorithms aim to compute similarity by combining classic metrics and handcrafted designs to capture data features or extract information.

\begin{figure}[h]
   \center
   \includegraphics[scale=0.52]{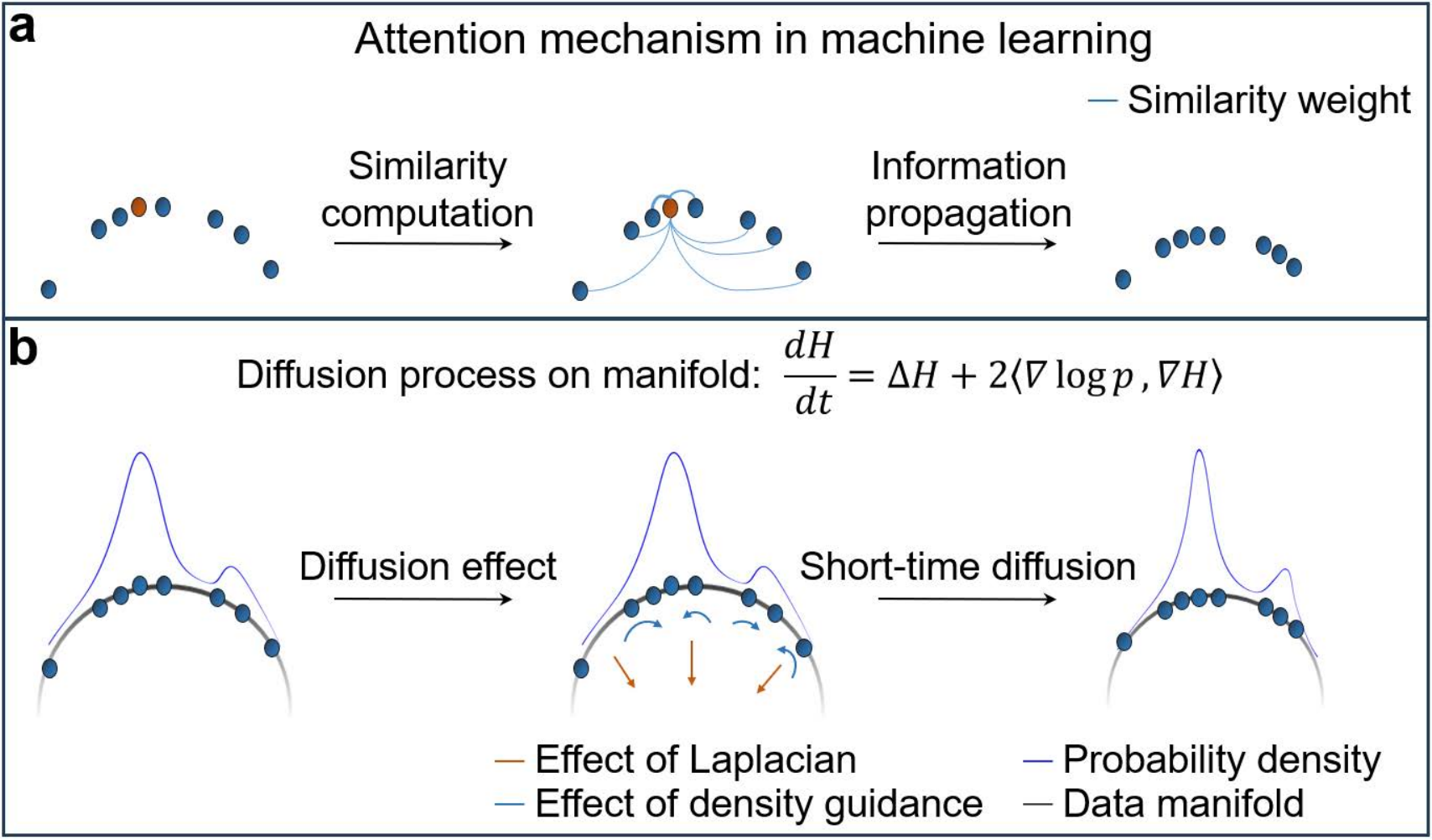}
   \caption{Illustration of the main idea. (a) Attention mechanism consists of two main steps: (1) computing the similarity between nodes (data points or tokens), followed by propagating node features to neighboring nodes, weighted by the similarities, and (2) updating the features of the nodes. (b) Illustration of a drift-diffusion process on the manifold where data reside. This process is driven by two main forces: density guidance, which encourages local concentration, and diffusion, which promotes global consistency of features. This study demonstrates that attention mechanism can be considered a first-order approximation of the drift-diffusion process on manifold, i.e., the short-time diffusion.}
   \label{fig:work}
\end{figure}

We can intuitively observe that attention mechanism also follows this principle of similarity computation to capture the inherent patterns in data. In this study, by exploring the common approach of similarity computation in classic machine learning algorithms, we explain how this technique is utilized in attention mechanism, thereby revealing its connections to classical machine learning algorithms. In addition, we illustrate that under certain assumptions like (1) the formulation of similarity can be decomposed into a transformation of a learnable metric function, a parameter on time scale and a softmax similarity computation; (2) the time-scale parameter is sufficiently small and (3) the manifold hypothesis holds, and there are sufficient data points sampled i.i.d. from a continuous distribution, attention mechanism for information propagation can be approximated by a drift-diffusion process on the manifold. Furthermore, we prove that, under certain assumptions, this process can be reformalized as a heat diffusion process under a new metric. This provides theoretical support for intuitively understanding the working principle of the self-attention mechanism through physical intuition (Figure \ref{fig:work}).

In assumption (1), if the learnable function is a general function rather than a metric function, we can still give a first-order analysis to describe the information propagation process of attention mechanism. Although this process does not generally correspond to a continuous dynamical system, it retains a similar interpretation to the metric version. That is to say, the zeroth-order effect is determined by the nearest data points, while the first-order effect relates to a drift-diffusion process at these points. The main difference between the two lies in how the nearest data points are defined.
 
From the perspective of similarity computation, we decompose the self-attention mechanism into an information propagation process based on handcrafted similarity computation and a learnable pseudo-metric. Inspired by this, we propose a modified attention mechanism by leveraging the concept of metric learning to enhance the ability to learn desired metrics more effectively. We refer to this modified attention mechanism as `metric-attention'.

In Section 2, we introduce the typical techniques of similarity computation used in many classic machine learning algorithms and clarify that attention mechanism also follows this principle. Additionally, we introduce the heat kernel approximation as it relates to attention mechanism. In Section 3, we explore the limit properties of attention mechanism for information propagation and prove that it can be regarded as a first-order approximation of a drift-diffusion equation under certain conditions, which can be further transformed into a heat equation. We also introduce a first-order analysis of information propagation in a generalized pseudo-metric setting. In Section 4, by interpreting attention mechanism as a combination of information propagation based on similarity computation and a learnable pseudo-metric, we propose a metric-attention mechanism to improve information propagation. Numerical experiments demonstrate its superior performance to the self-attention mechanism on various examples. Finally, we conclude this work and discuss its implications.

\section{From similarity computation to attention mechanism}
\subsection{Similarity computation}
Similarity computation is a set of engineering practices to generate similarity measures between data points (or graph nodes). Different methods for similarity computation have been developed. Most of them involve one, two, or three of the following three components: initializing similarity, strengthening similarity, and normalizing similarity.

\subsubsection{Initializing similarity}
Initializing similarity is the first step for further computations and information extraction. The point-to-point similarity is typically generated using a binary function $D(\cdot,\cdot)$, where $D$ is often a simple transformation based on metrics, inner products, or topological structures.
 
A natural idea is that the closer two data points are, the higher their similarity. Therefore, the distance between two data points is often transformed into a similarity measure through monotonically decreasing mappings. Building on previous work, we have derived the following metric-based function $D_{t,c}$: 
 
\begin{Definition}[Metric-based similarity generation]
Given a metric space $(\mathcal{M},d)$, we define the similarity function on $\mathcal{M}$ as:
\begin{align*}
   D_{c,t}(x,y)=c(x)-\rm{sign}(t)d(x,y)^t
\end{align*}
where $c$ is a specified function and $t$ is a hyperparameter.
\end{Definition}
 
Bilinear functions are also commonly used to initialize similarity.
\begin{Definition}[$QK$-dot product]
Given $m\times n$ matrices $Q,K$ and vectors $x,y\in \mathbb{R}^n$, where $x$ and $y$ are column vectors, we define $QK$-dot product of $x$ and $y$ as follows:
\begin{align*}
D(x,y)=x\cdot_{QK} y=Qx\cdot Ky=x^T (Q^TK) y
\end{align*}
\end{Definition}
Before the inner product mapping, people may apply a non-linear transformation to the original features, which is equivalent to defining similarity using a certain kernel function $\mathbf{K}(\cdot,\cdot)$.

Additionally, similarity may be defined using a local combination, which reflects how a data point or node is represented by a combination of its neighbors.
\begin{Definition}[Local combination similarity]
Given a set of data points $\{v_i\}\subset \mathbb{R}^n$ and their adjacency relationships, 
we calculate $\omega_{ij}$ to satisfy:
\begin{align*}
    v_i=\sum_{v_j\in \mathcal{N}(i)} \omega_{ij}v_j
\end{align*}
where $\mathcal{N}(i)$ is the set of neighbors of $v_i$. The local combination similarity between $v_i$ and $v_j$ is $\omega_{ij}$, denoted by ${\rm Lc}(v_i,v_j)$.
\end{Definition}

In addition to feature-based similarity initialization methods, topology-based methods have also been defined:
\begin{Definition}[$k$-th order adjacency similarity]
Given a graph $G(V=\{v_i\},E=\{(v_{a_j},v_{b_j})\})$, with adjacency matrix $A$, we define the $k$-th order adjacency similarity of $v_i$ and $v_j$ as $(A^k)_{ij}$.
\end{Definition}

\subsubsection{Strengthening similarity}
The purpose of strengthening similarity is to make two data points that are relatively similar become even more similar. It can work in conjunction with the normalization process, by strengthening similarities and decreasing weaker ones to aid the aggregating process.
\begin{Definition}[$r$-Inflation]
Given the hyperparameter $r$, we define the inflation operator $\Gamma_r$:
\begin{align*}
   \Gamma_r:M_{m\times n}&\to M_{m\times n}\\
   \left(\Gamma_r(M)\right)_{ij}&={\rm sign}(r)M_{ij}^r
\end{align*}
\end{Definition}
 
\begin{Definition}[Exponential Inflation]
Given a vector $\epsilon=(\epsilon_1,\cdots,\epsilon_n)$ of size n, we define the exponential inflation operator $\Gamma_\epsilon^*$:
\begin{align*}
   \Gamma_\epsilon^*:M_{m\times n}&\to M_{m\times n}\\
   (\Gamma_\epsilon^*(M))_{ij}&=\exp\left(\frac{M_{ij}}{\epsilon_i}\right)
\end{align*}
\end{Definition}
 
\subsubsection{Normalizing similarity}
The purpose of normalizing similarity is to transform the similarity matrix into a desired form, which is often related to the modeling of the data. Given a similarity matrix $S$, where each element is positive, four common normalization operations are typically used.
\begin{Definition}[Row normalization]
\begin{align*}
   N_r:M_{m\times n}&\to M_{m\times n}\\
   (N_r(S)_{ij}&=\frac{S_{ij}}{\sum_k S_{ik}}
\end{align*}
\end{Definition}
 
\begin{Definition}[Column normalization]
\begin{align*}
   N_c:M_{m\times n}&\to M_{m\times n}\\
   (N_c(S)_{ij}&=\frac{S_{ij}}{\sum_k S_{kj}}
\end{align*}
\end{Definition}
 
\begin{Definition}[Two-side normalization]
\begin{align*}
   N_2:M_{m\times n}&\to M_{m\times n}\\
   (N_2(M)_{ij}&=\frac{M_{ij}}{\sum_k M_{ik}\sum_k M_{jk}}
\end{align*}
\end{Definition}
 
\begin{Definition}[Global normalization]
\begin{align*}
   N_g:M_{m\times n}&\to M_{m\times n}\\
   (N_g(M)_{ij}&=\frac{M_{ij}}{\sum_{k,l} M_{kl}}
\end{align*}
\end{Definition}

\subsection{Similarity computation in machine learning}
Here, we introduce the applications of similarity computation in manifold learning, clustering, supervised learning, and attention mechanism in neural networks, highlighting the distinct common characteristics shared by attention mechanism and traditional algorithms.
 
\subsubsection{Similarity computation in manifold learning}
Manifold learning refers to dimensionality reduction. Due to the curse of dimensionality, high-dimensional data are often difficult to handle and cannot be visualized. Therefore, dimensionality reduction methods are employed to preprocess the data. The problem statement of manifold learning is as follows: given the high-dimensional data points $\{x_i, i=1,\cdots,N\}\subset \mathbb{R}^n$, assuming that they are distributed on a low-dimensional manifold, how to find their corresponding data points $\{y_i,i=1,\cdots,N\}$ in a low-dimensional space $\mathbb{R}^l$, where $l\ll n$, such that the topology, distances or densities of data points on the underlying manifold is preserved as much as possible?
 
How can we maintain the main structure of data during the process of dimensionality reduction? Prior work often adopts the principle of similarity (dissimilarity or distance) preservation. First, the coordinates of data points in the low-dimensional space are determined using some initialization method. Then, the coordinates are optimized so that the pairwise similarities in the low-dimensional space closely approximate those in the original data. The key difference among most approaches lies in the different similarity computation techniques they employ. To show this, we summarize those used in several classical algorithms including MDS \citep{kruskal1978multidimensional}, Isomap \citep{balasubramanian2002isomap}, LLE \citep{roweis2000nonlinear}, Laplacian eigenmaps \citep{belkin2003laplacian}, Diffusion map \citep{coifman2006diffusion}, SNE \citep{hinton2002stochastic}, t-SNE \citep{van2008visualizing}, UMAP \citep{mcinnes2018umap} (Table \ref{tab:manifold_learning}). These methods employ hand-crafted similarity computation techniques.

\begin{table}[h]
\caption{Summary of typical techniques for computing similarity in manifold learning methods.}
\centering
\begin{tabular}{cccccc}
\toprule
\textbf{Method} & \begin{tabular}[c]{@{}c@{}}\textbf{Similarity} \\\textbf{initialization}\end{tabular} & \begin{tabular}[c]{@{}c@{}}\textbf{Similarity} \\\textbf{strengthen}\end{tabular} & \textbf{Normalization}  &   \\
\midrule
   MDS &    $D_{1,0}$                  &     None                  &   None            &                \\
   Isomap &       $D_{1,0}$ (Graph)                &         None              &        None       &                \\
   LLE  & Lc &None &None & \\
   Laplacian eigenmaps & $D_{2,0}$ & $\Gamma_{\epsilon}^*$ & None \\
   Diffusion map & $D_{2,0}$ & $\Gamma_{\epsilon}^*$ & $N_r\circ N_2$\\
   SNE & $D_{2,0}$ & $\Gamma_{\epsilon}^*$ & $N_r$& \\
   t-SNE &    $D_{2,0}$ and $D_{2,-1}$                  &       $\Gamma_\epsilon^*$ and $\Gamma_{-1}$                 &   $N_g$ and $N(W)=1/2W+1/2W^T$           &                \\
    UMAP  & $D_{1,c}$ and $D_{c,-1}$                      &     $\Gamma_{\epsilon}^*$ and $\Gamma_{-1}$                  &   $N(W)=W+W^T-W\odot W^T$            &                \\
\bottomrule
\end{tabular}
\label{tab:manifold_learning}
\end{table}

\subsubsection{Similarity computation in clustering}
\paragraph{Fuzzy c-means clustering and k-means clustering}
Fuzzy c-means clustering categorizes data points into several classes such that data points within each class have high similarity. This is achieved by alternately calculating similarity and updating class representatives based on similarity. To be specific, suppose we have data points $\{x_1,\cdots,x_N\}\subset\mathbb{R}^n $ and class center $\{c_1,\cdots,c_m\} \subset \mathbb{R}^n$:
\begin{itemize}
   \item Calculate the similarity between data points and class center:
   \begin{align*}
   (D_{\frac{-2}{m-1},0})_{ij}&=\Vert c_i-x_j\Vert^{\frac{-2}{m-1}}\\
       S&=N_r\circ N_c(D_{\frac{-2}{m-1},0})
   \end{align*}
   where $m$ is a hyper-parameter.
   \item Renew class center:
   \begin{align*}
       c_i=\sum_j S_{ij}x_j
   \end{align*}
\end{itemize}
Repeat the above process until convergence. The $c_i$ represents the representative of class $i$ and $ (N_c\circ D_{\frac{-2}{m-1},0})_{ij}$ represent the probability that data point $j$ belongs to class $i$. By taking the limit $\frac{-2}{m-1}\to-\infty$ in the fuzzy c-means algorithm, we obtain the k-means algorithm.
 
\paragraph{Markov clustering algorithm (MCL)} MCL is an unsupervised graph clustering algorithm. The input is a (weighted) graph, and the output is a clustering of nodes. Here, we describe the MCL process using the concepts of information passing and feature representation. MCL iteratively calculates:
\begin{align*}
   H^{k+1}=S^{(k)}\cdot I
\end{align*}
where $H^{k+1}$ represents the clustering result of the algorithm at the $k$-th iteration, $I$ is the unit matrix which represents the feature matrix of one-hot vector, and the similarity matrix is computed as:
\begin{align*}
S^{(k)}=N_r \circ \Gamma_2 \circ M_2 (H^k)    
\end{align*}
where $M_2(A):=A^2$, indicating the topological similarity of second order, $\Gamma_2$ is element-wise squaring, and $N_r$ represents the normalization performed on each row. $H^1$ equals the input weight matrix.
This process can be seen as computing node similarity iteratively and updating node features based on this similarity. Once the process converges, the features of each node are used to classify the nodes.
 
\subsubsection{Similarity computation in supervised learning}
\paragraph{K-nearest neighbors algorithm (KNN)} KNN is a supervised classification algorithm that utilizes adjacency information \citep{fix1985discriminatory}. Given a set of data points $\{(x_i, y_i),i=1,\cdots,N\}$, where $y_i\in\{0,1\}$ represents the label, the predicted label $\hat{y}$ for a test data point $\hat{x}$ is computed as $\hat{y}=\boldsymbol{s}^T \boldsymbol{y}$, where $\boldsymbol{y}=(y_1,\cdots,y_N)^T$, and the similarity vector $\boldsymbol{s}=(s_1,\cdots,s_N)^T$ is defined by:
\begin{align*}
   \boldsymbol{s}=N_c(A)
\end{align*}
where $A=(A_i,\cdots,A_N)^T$. $A_{i}=1$ if $x_i$ is one of the k-nearest neighbors of $\hat{x}$; otherwise, $A_{i}=0$.
 
\paragraph{Support vector machine (SVM)}
SVM is a supervised binary classification algorithm \citep{cortes1995support}. Given a dataset $\{(x_i, y_i),i=1,\cdots,N\}$, where $y_i \in \{1,-1\}$, and a kernel function $\mathbf{K}(\cdot,\cdot)$. SVM first computes the coefficients $\alpha_i$ and the bias term $b$ corresponding to the supporting plane. For a new sample $\hat{x}$, the predicted label is computed as $\hat{y} = \boldsymbol{s}^T\boldsymbol{y} + b$, where $\boldsymbol{y}=(y_1,\cdots,y_N)^T$ and $\boldsymbol{s}=(s_1,\cdots,s_N)^T$ is the similarity vector defined as:
\begin{align*}
   \boldsymbol{s}_i=\alpha_i \mathbf{K}(\hat{x},x_i)
\end{align*}
If $\hat{y}>0$, then $\hat{x}$ is classified into the class of $+1$; otherwise, it is classified into the class of $-1$.

\subsubsection{Similarity computation in attention mechanism for information propagation}
\paragraph{Neural networks}
Information propagation modules are highly prevalent in neural network architectures, particularly in graph neural networks and Transformers. These modules are typically followed by linear transformations and nonlinear activations, which together form the core of these architectures. An information propagation module can be expressed as:
\begin{align*}
   H^{k+1}=S^{(k)}H^{k}
\end{align*}
where $H^k$ contains the features of each node in the $k-$th layer, $S^{(k)}$ is the similarity matrix in which $S^{(k)}_{ij}$ represents the similarity between the $i$-th and $j$-th data points in the $k$-th layer. Different neural networks adopt various methods for similarity computation. For example, 
\begin{itemize}
   \item Graph convolutional networks: $S={\rm Normalized}(A+I)$, where $A$ is the adjacency matrix.
   \item Diffusionnet \citep{sharp2022diffusionnet}: $S=\exp(tL)$, where $t$ is a parameter and $L$ is the discrete Laplacian matrix.
   \item Transformer and graph attention network: Similarity between data points is initially computed pairwise through a learnable pseudo-metric function $f_\theta(x_i,x_j)$. The resulting similarity matrix is then subjected to exponential scaling and row normalization to yield $S$:
   \begin{align*}
       S_{ij}=\frac{\exp \left(-f_\theta(x_i,x_j)\right)}{\sum_k \exp\left(-f_\theta(x_i,x_k)\right)}
   \end{align*}
\end{itemize}
 
We observe that attention mechanism utilizes similarity computation and information propagation, that are core components of classical algorithms. The key difference is that classical algorithms often rely on manually designed similarity computation methods, which limits their applicability. In contrast, attention mechanism incorporates learnable parameters, which makes them adaptive and suitable for more general network architectures. Moreover, this formulation closely resembles the heat kernel approximation, enabling an analysis of the properties of the attention mechanism from this perspective.
 
\subsection{Approximate Laplacian-Beltrami operator by heat kernel}
\label{heat_kerbel_approx}
The heat kernel is deeply connected to the Laplacian-Beltrami operator, which is a fundamental tool in differential geometry and mathematical physics for studying the geometric properties of surfaces and manifolds. This operator generalizes the Laplacian from Euclidean spaces to general manifolds. Informally, it contains all the information of the Riemannian manifold \citep{bronstein2010scale}. Here, we focus on the 0-form Laplacian operator, as it is the most commonly studied and has the most direct connection to attention mechanism.

In the following, we introduce the method to approximate the Laplacian-Beltrami operator using the heat kernel. Given the data points $\{x_i\}_{i=1}^N$ and the heat kernel $h_\epsilon(x,y)=\exp{\left(-\frac{\lVert x-y\rVert^2}{2\epsilon}\right)}$, we define the weight matrix $W$ of $N\times N$ by heat kernel $h_\epsilon$ and the diagonal matrix $D$: 
\begin{align*}
   W_{ij}=h_\epsilon(x_i,x_j)=\exp{\left(-\frac{\lVert x_i-x_j\rVert^2}{2\epsilon}\right)},\quad
   D_{ii}=\sum_{j=1}^n W_{ij}=\sum_j \exp{\left(-\frac{\lVert x_i-x_j\rVert^2}{2\epsilon}\right)}
\end{align*}
The negative defined graph Laplacian for data points $\{x_i\}_{i=1}^N$ is defined as $L=D^{-1}W-I$, where
\begin{align*}
L_{ij}=\frac{\exp{\left(-\frac{\lVert x_i-x_j\rVert^2}{2\epsilon}\right)}}{\sum_k \exp{\left(-\frac{\lVert x_i-x_k\rVert^2}{2\epsilon}\right)}}-\delta_{ij}
\label{heatkernelapprox}
\end{align*}
where $\delta_{ij}=1$ if $i=j$ and $\delta_{ij}=0$ otherwise. Given that the data points $\{x_i\}_{i=1}^N$ are uniformly distributed on a manifold, it has been shown that the graph Laplacian will converge to the Laplacian of the manifold as $\epsilon\to 0$ and $N\to \infty$. The following theorem implies that a heat kernel can approximate the Laplacian of a Riemannian manifold and provides the convergence rate of this approximation.
\begin{Theorem}[Naive heat kernel approximator \citep{singer2006graph}]
Suppose $\{x_i\}_{i=1}^N$ are i.i.d. sampled from the uniform measure on a compact Riemannian manifold, then
\begin{align*}
    \frac{1}{\epsilon}\sum_{j=1}^N L_{ij}f(x_j)=\frac{1}{2}\Delta f(x_i)+\mathcal{O}\left(\frac{1}{N^{1/2}\epsilon^{1/2+d/4}},\epsilon\right)
\end{align*}
where $f(\cdot)$ is a smooth function, $\Delta$ is the (negative defined) Laplacian-Beltrami operator of the Riemannian manifold, and $\mathcal{O}$ represents the big O notation.
\end{Theorem}
 
When the data are not sampled from a uniform distribution, \citet{coifman2006diffusion} offered a general Laplacian approximator. If one still opts to use the Naive heat kernel approximator, it will result in an additional component. The following lemma describes the influence of distribution, and it is the key to depict the limit property of attention mechanism:
\begin{Theorem}
[Deviation of naive heat kernel approximator \citep{coifman2006diffusion}]
Suppose the data points $\{x_i\}_{i=1}^N$ are i.i.d. sampled from a distribution on a compact Riemannian manifold with density function $p(x)$, then
\begin{align*}
    \frac{1}{\epsilon}\sum_{j=1}^N  L_{ij}f(x_j)\to \frac{1}{2}\left(\Delta f(x_i)+2\left\langle \frac{\nabla p}{p}(x_i), \nabla f(x_i)\right\rangle\right)
\end{align*}
\end{Theorem}

\section{Limit properties of attention mechanism for information propagation}
Due to the analogous formulations of heat kernel approximation and information propagation of attention mechanism, we can analyze the asymptotic properties of attention mechanism by analogy. To start with, we will show a direct analysis for the metric setting of the learnable pseudo-metric, which corresponds to an elegant continuous dynamical system. This special setting is helpful for intuitively understanding attention mechanism by physical intuition. We further generalize this analysis to the situation where the learnable pseudo-metric function does not satisfy the metric setting. The primary difference between these two settings lies in how neighbors are defined. 

\subsection{Assumptions}
\label{Assumptions}
\paragraph{Network structure formulation}
We examine the attention coefficients computation and information propagation steps in Transformer and graph attention network. Specifically, we pay attention to the following steps:
\begin{align*}
   H^{\rm new}=SH^{\rm old}
\end{align*}
where
$S_{ij}=\frac{\exp \big(-f_\theta(H^{\rm old}_i,H^{\rm old}_j)\big)}{\sum_k \exp\big(-f_\theta(H^{\rm old}_i,H^{\rm old}_k)\big)}$. In the Transformer block, $f_\theta(x,y)=-x^T(Q^TK)y.$ In the graph attention network, $f_\theta(x,y)=-\sigma(\alpha_1^TPx-\alpha_2^TPy)$, where $\alpha_i\,(i=1,2)$ are learnable vectors, $P$ is a learnable matrix and $\sigma$ is an activation function.
 
We reformulate the similarity matrix of attention mechanism $S$ as $S_\epsilon$:
\begin{align*}
    S_{\epsilon,ij}=\frac{\exp \big(-\frac{f_\theta(H^{\rm old}_i,H^{\rm old}_j)}{2\epsilon}\big)}{\sum_k \exp\big(-\frac{f_\theta(H^{\rm old}_i,H^{\rm old}_k)}{2\epsilon}\big)}
\end{align*}
where $\epsilon/2$ represents the time scale. In attention mechanism, the updated representation is given by $H^{\rm new}=S_{\epsilon}H^{\rm old}$, where $\epsilon=\frac{1}{2}$.
 
By this formulation, we suppose that:
\begin{itemize}
   \item \textbf{Assumption 1}: $\epsilon$ is sufficiently small.
   \item \textbf{Assumption 2}: $\sqrt{c+f_\theta(\cdot,\cdot)}$ is a distance function $d_\theta$ induced by a Riemannian metric $g_\theta$, where $c$ is a constant.
\end{itemize}
\begin{Example}
In Transformer, if $Q=K=I$, then $f_\theta(x,y)=-x\cdot y.$ Given that the data is distributed on the unit sphere, we have:
\begin{align*}
   f_\theta(x,y)=-2+\Vert x-y\Vert^2_2
\end{align*}
Generally, if $Q=K$, then $f_\theta(x,y)=-x^T Q^TQ y.$ Given that the data is distributed on the ellipsoid $\{x\in \mathbb{R}^d:x^T Q^TQ x=1\}$, we have:
\begin{align*}
   f_\theta(x,y)=-2+\Vert x-y\Vert_Q^2
\end{align*}
where $\Vert x-y\Vert_Q^2=(x-y)^TQ^TQ(x-y)$.
\end{Example}
 
\begin{Example}
In graph attention network, if $\alpha_1=-\alpha_2$ and the activation is symmetric about the origin, then:
\begin{align*}
   f_\theta(x,y)=-\sigma\big(\alpha_1P(x-y)\big)
\end{align*}
also satisfies symmetry. If the data is distributed on a 1-d manifold and $\sigma(x)=-x^2$, then $f_\theta(x,y)$ may satisfy the properties described above.
\end{Example}
 
Our first assumption primarily facilitates the application of Theorem 2, setting the stage for the convergence of the information propagation in the attention mechanism to a partial differential operator. Although idealized, our second assumption is relatively reasonable in the context of neural network configurations and similarity computations, offering critical insights. As demonstrated in the previous examples, attention mechanism settings may meet some assumptions under certain conditions.
 
\paragraph{Assumption 3 (Data distribution assumption)}
The final assumption relates to the manifold hypothesis, which posits that meaningful high-dimensional data in real life often lies on an intrinsic low-dimensional manifold, which is the theoretical basis for many machine learning algorithms \citep{roweis2000nonlinear,coifman2006diffusion,WOLD198737}. This hypothesis has been validated in numerous cases, such as the MNIST handwritten digit database \citep{deng2012mnist}, and has been widely adopted. Based on this hypothesis, many algorithms have been developed for manifold fitting \citep{yao2024manifold,yao2019manifold} and manifold learning \citep{roweis2000nonlinear,coifman2006diffusion}, among others.
 
Following the manifold hypothesis, we assume that the data are distributed on a (compact, connected) Riemannian manifold $\mathcal{M}$, and they are i.i.d. sampled from a random variable $X$ on $\mathcal{M}$. This random variable has a density $p(x)$ with respect to the volume element $\omega$ of $(\mathcal{M},g_\theta)$:
\begin{align*}
   \mathbb{P}(X\in A)=\int_A p(x) d\omega,\, A\in \mathcal{B}(\mathcal{M})
\end{align*}
where $\mathcal{B}(\mathcal{M})$ refers to the Borel set, $g_\theta$ is the Riemannian metric in Assumption 2. Through an initial embedding $H^{\rm old}$, we observe the data in Euclidean space:
\begin{align*}
   H^{\rm old}:\mathcal{M}\hookrightarrow \mathbb{R}^n
\end{align*}
We also assume that the dataset is sufficiently large ($N\to \infty$). Thus, it is natural to adopt a continuous perspective.

\subsection{Attention limit operator.}
\begin{Theorem}
[Limit of attention mechanism for information propagation]
\label{attention_limit_pde}
Suppose the assumptions are satisfied, then the dynamics of the feature of each data point in attention mechanism for information propagation is a first-order approximator (with respect to $\epsilon$) of the PDE:
\begin{align*}
    &\frac{dH}{dt}=\Delta_{g_\theta} H+2 \left\langle\frac{\nabla p}{p}, \nabla H\right\rangle\\
    &H(x,0)=H^{\rm old}(x)
\end{align*}
where $\Delta_{g_\theta}$ is the Laplacian-Beltrami operator of manifold $\mathcal{M}$ with the Riemannian metric given by Neural network (Assumption 2) and $p$ is the density function (Assumption 3).
\end{Theorem}
Based on this theorem, we define the attention limit operator ${\rm At}_{g,p}$ by $\Delta_g+2  \left\langle\frac{\nabla p}{p}, \nabla\cdot\right\rangle$, or equivalently $\Delta_g+2  \left\langle\nabla\log p, \nabla_g\cdot\right\rangle$, which characterizes the limit increment of features influenced by the information propagation of attention mechanism.
 
The attention limit operator reflects the combined information propagation effect of diffusion and density-guided flow (Figure \ref{fig:work}). The ${\rm At}_{g,p}$ limit operator contains two parts: the Laplacian term $\Delta$ and the particle drift $\left\langle\frac{\nabla p}{p}, \nabla\cdot\right\rangle$ term. The Laplacian term represents heat diffusion, indicating heat diffuses uniformly in all directions within a homogeneous medium, serving to make features tend towards consistency. The particle drift term represents particles always moving in the direction of the steepest change in probability density rather than along contour lines, suggesting that density guides the flow of information. Additionally, this PDE can be translated into its stochastic differential equation (SDE) counterpart, which offers a similar interpretation.
 
Using this theorem, we can analyze the impact of metric scaling on the information propagation rate. If $g_2=\frac{g_1}{m^2}$, $m$ is a constant bigger than $1$, then:
\begin{align*}
   \Delta_{g_2}=m^2 \Delta_{g_1}
\end{align*}
This implies that the rate of information propagation through diffusion can be increased by compressing distances. The density-guided flow also varies with different metrics. Therefore, neural networks can modulate the rate of information propagation by learning different metrics. 
 
\begin{Theorem}[Heat diffusion formulation]
   If the dimension of the manifold $\mathcal{M}$ is $n\neq2$, then there exists a metric $\tilde{g}$ such that the ${\rm At}_{g,p}$ operator is equivalent to a Laplacian-like operator of $\tilde{g}$:
   \begin{align*}
       \Delta_g+2\left\langle\frac{\nabla p}{p}, \nabla \cdot \right\rangle=f\Delta_{\tilde{g}}
   \end{align*}
   where $f=p^{4/n-2}$. Therefore, attention mechanism for information propagation can be described by the heat diffusion equation:
   \begin{align*}
       \frac{dH}{dt}=f\Delta_{\tilde{g}}H
   \end{align*}
\end{Theorem}
 
This theorem allows us to interpret the dynamics of the attention operator by heat diffusion, where $f$ can be regarded as the specific heat capacity. To be specific, the heat diffusion on the manifold $(\mathcal{M},\tilde{g})$ with specific heat capacity $c$, thermal conductivity $k$ and material density $\rho$ follows the dynamics (Appendix \ref{proof}):
\begin{align*}
   \frac{du}{dt}=\frac{1}{c\rho}\nabla \cdot(k\nabla u)
\end{align*}
where $u(x,t)$ is the temperature of $x$ at time $t$. When $k=1,\rho=1$, and $c=f^{-1}$, this dynamic is the same as the dynamic of attention mechanism.
 
In this case, attention mechanism essentially learns a new metric on the manifold and the data features undergo the transformation process like heat conduction under this new metric. In this view, neural networks adjust the metric automatically so that features evolve favorably for tasks. For example, in classification tasks, we desire faster heat conduction within class regions, leading to quicker feature averaging; and between-class separation regions undergo slower heat conduction, allowing for distinct features between classes. 
 
\paragraph{Stationary function (Clustering tendency).}
By comparing this PDE with the standard heat equation, we explore the equilibrium states of the equation.
\begin{align*}
   \text{Attention limit dynamics:}\,\frac{dH}{dt}&=f\Delta H\\
   \text{Standard heat diffusion:}\,\frac{dH}{dt}&=\Delta H
\end{align*}

Since the information propagation of vanilla self-attention mechanism under the metric setting represents a first-order approximation of the PDE, we assume that $f$ is a fixed function. Due to the positivity of $f$, the two equations above share the same stable states: any function that is stable under the attention limit dynamics is also stable under the standard heat equation, and vice versa. Hodge theory provides insight into the relationship between cohomology classes and harmonic equations: the dimension of harmonic $n$-forms equals the dimension of the $n$-th cohomology group. We have:
\begin{align*}
   \dim\{f:\Delta f=0\}=\dim(H^0)
\end{align*}

Thus, if a manifold is connected, the harmonic equations on it admit only constant solutions. (Such inferences are not only valid on manifolds. Similar conclusions exist in graph-based Hodge theory). Therefore, the attention limit dynamic can only stabilize at constant functions, which implies a clustering of features. In practice, this clustering phenomenon does occur when there are too many attention blocks without skip connections \citep{dong2021attention}. Moreover, according to spectral theory, the number of near-zero eigenvalues reflects the number of clusters that may form before merging into a single cluster. This property provides insight into the finite-time clustering behavior of the dynamical system.

\subsection{Multi-head attention}
Multi-head attention combines multiple attention blocks. A $k$-head attention mechanism is defined as:
\begin{align*}
    &H_i=S_iH^{\rm old},i=1,\cdots,k\\
    &H^{\rm new}=[H_1V_1,\cdots,H_kV_k][W_1,\cdots,W_k]^T
\end{align*}
where $S_i$ is the similarity matrix for the $i$-th head, $V_i$ is the value matrix of the $i$-th head, $W=[W_1,\cdots,W_k]$ is a learnable matrix. Reformalize the formula we have:
\begin{align*}
    H^{\rm new}=\sum_i^k S_i H^{\rm old}\tilde{W_i}
\end{align*}
where $\tilde{W}_i=V_iW_i^T$. Consequently, when assumptions \ref{Assumptions} are satisfied, a multi-head self-attention block can be viewed as a first-order approximator of a combination of $k$ PDEs:
\begin{align*}
    &\frac{dH_i}{dt}=\Delta_{g_i} H+2 \left\langle\frac{\nabla p_i}{p_i}, \nabla H\right\rangle\\
    &H_i(x,0)=H^{\rm old}(x)\\
    &H^{\rm new}=\sum_i H_i W_i
\end{align*}
where $g_i$ is the Riemannian metric learned by the $i$-th attention block and $p_i$ is the density function of the random variable, from which data are sampled, with respect to $g_i$. Compared to the single-head attention mechanism, multi-head attention simultaneously learns multiple pseudo-metrics for information aggregation and combines the aggregated information. This architecture may reduce the difficulty of learning meaningful pseudo-metrics, thereby demonstrating a stronger capability to extract information.

\subsection{An analysis for general pseudo-metric setting}
\label{general pseudo-metric setting}
Before presenting our analysis for the information propagation process of general pseudo-metrics in attention mechanism, we provide an explanation of Theorem \ref{attention_limit_pde} to strengthen our understanding of the role of the metric:
\begin{align*}
    &H^{\rm new}(x) = \underbrace{H^{\rm old}(x)}_{\rm Zeroth-order}+ \underbrace{\frac{\epsilon}{2}\left(\Delta_{g_\theta} H^{\rm old}(x)+2 \left\langle\frac{\nabla p}{p}(x), \nabla H^{\rm old}(x)\right\rangle\right)}_{\rm First-order} + {\rm Higher\,order}
\end{align*}
The zeroth-order term represents the information at the data point $x$ itself, which can be considered the `nearest' point to $x$ as measured by the metric: since a distance function is positive definite by definition, the only data point that is nearest to $x$ is itself. Similarly, the first-order term comes from the the drift-diffusion process at the nearest data point of $x$ measured by the metric. Inspired by this, if we use a general pseudo-metric function which may not satisfy the conditions of a metric, the zeroth-order term should correspond to the information at the nearest data points of $x$ and the first-order term should be related to a drift-diffusion process at the nearest data points of $x$, where the nearest data points is define by pseudo-metric $f_\theta(x,y)$. The main difference is that a metric function $d$ should satisfy the following three conditions:
\begin{itemize}
    \item $d(x,y)\ge 0$ and $d(x,x)=0 \iff x=y$ 
    \item $d(x,y)=d(y,x)$
    \item $d(x,y)\le d(x,z)+ d(z,y)$
\end{itemize}

Therefore, for a metric function, the only data point nearest to $x$ should be itself, while for general pseudo-metric functions, the nearest data points may not be $x$ itself and not be unique. 

\begin{Example}
In Transformer, given the three conditions hold: (1) $-Q^TK= P^T {\rm diag}(a_1,\cdots,a_n)P$, where $P\in SO(n)$; (2) $f_\theta(x,y)=\sum a_ix'_iy'_i$, where $x'=Px$ and $y'=Py$; (3) $x$ and $y$ lie on the unit sphere and there exists $a_ix_i'\neq0$, we obtain the following result (Appendix \ref{proof_of_general pseudo-metric_setting}): 
\begin{align*}
   {\rm argmin}_yf_\theta(x,y)=P^{T}\left[\frac{a_1x'_1}{\sqrt{\sum a_i^2x_i'^2}},\cdots,\frac{a_nx'_n}{\sqrt{\sum a_i^2x_i'^2}}\right]^T
\end{align*}
Generally, if $Q^TK$  is non-degenerate and $x,y$ lie on an ellipsoid, $f_\theta(x,y)$ has a unique minimizer $y$. 
\end{Example}

Denote ${\rm argmin}_yf_\theta(x,y)$ as $A_x$. According to our previous analysis, the zeorth-order effect should be an average of information in $A_x$ and the first-order effect should be related to a drift-diffusion process. 
\begin{Theorem}[Informal]
\label{general_thm_y'}
If $A_x=\{y'\}$ and certain regularity conditions hold (Appendix \ref{proof_of_general pseudo-metric_setting}), we have:
\begin{align*}
    H^{\rm new}(x) = H^{\rm old}(y')+\epsilon {\rm At}_{f_\theta,p}H^{\rm old}(y') + {\rm Higher\,order}
\end{align*}
where ${\rm At}_{f_\theta,p}$ is a second-order partial differential operator related to the pseudo-metric $f_\theta$ and the sampling density $p$.
\end{Theorem}
We prove this theorem, along with a generalized version in which $A_x$ is a manifold in Appendix \ref{proof_of_general pseudo-metric_setting}. This first-order analysis demonstrates that using a general function as the pseudo-metric provides a novel approach for defining nearest points, thereby effectively establishing a new topology. Since the zeroth-order term may differ from the original feature, this approach could offer a more flexible and powerful tool for information propagation. However, it also implies that the training process might be more challenging since this information propagation process does not necessarily correspond to a continuous dynamical system.

\section{From self-attention to metric-attention} 
\subsection{Metric-attention}
\label{Metric-attnetion}
As demonstrated above, attention mechanism is essentially a combination of an information propagation process based on a handcrafted similarity computation and a learnable pseudo-metric $f_\theta(\cdot,\cdot)$. We expect that neural networks can learn beneficial pseudo-metrics from data. However, the self-attention mechanism formulates the pseudo-metric as:
\begin{align*}
    f_\theta(x,y)=-x^TQ^TKy
\end{align*}
which can be interpreted as performing a linear transformation on features followed by dot product.These overly simplistic pseudo-metrics may struggle to capture complex similarity relationships, and intuitively, we would prefer them to correspond to continuous dynamical systems for easier training and better generalization. Therefore, we draw on the concept of metric learning \citep{yu2016deep,hu2014discriminative} and propose a modified attention mechanism called metric-attention mechanism, in which $f_\theta(x,y)=\lVert \tilde{f}_\theta(x)-\tilde{f}_\theta(y)\rVert_2^2$, where $\tilde{f}_\theta$ is a learnable function. The metric-attention information propagation mechanism is detailed in Algorithm \ref{alg:metric-attention}:

\begin{algorithm}
\caption{Metric-attention information propagation mechanism}
\label{alg:metric-attention}
\begin{algorithmic}[1]
\REQUIRE Data matrix $H^{\rm old}= [h_1,h_2,\cdots,h_N]^T$, a neural network $\tilde{f}_\theta$.
\STATE Calculate similarity matrix $S$,  
\begin{align*}
 S_{ij}=\frac{\exp\left(-\lVert \tilde{f}_\theta(h_i)-\tilde{f}_\theta(h_j)\rVert^2_2\right)}{\sum_k \exp\left(-\lVert \tilde{f}_\theta(h_i)-\tilde{f}_\theta(h_k)\rVert_2^2\right)} 
\end{align*}

\STATE Calculate $H^{\rm new} = SH^{\rm old}$

\RETURN $H^{\rm new}$
\end{algorithmic}
\end{algorithm}

 Note that when $\tilde{f}_\theta(x)$ adopts a linear transformation, the model reduces to the L2 self-attention mechanism \citep{kim2021lipschitz}. Generally, the network $\tilde{f}_\theta$ requires sufficient parameters to capture complex relationships while remaining easy to train. To achieve this, we suggest a single-hidden-layer MLP with a residual connection, which strikes a balance between these two factors, and this setting is adopted in our experiments for demonstration. In addition, this information propagation mechanism can be easily extended to graph data by setting the similarity between non-adjacent nodes to zero. 

\subsection{Experiments}
To evaluate the effectiveness of the metric-attention mechanism while minimizing the influence of other modules, we construct an information propagation network (IPN) by sequentially passing the input features through a linear transformation layer, a series of attention-based information propagation modules, and another linear transformation layer. For classification tasks, we employ a softmax classifier to classify features. We choose the self-attention mechanism, L2 self-attention mechanism, and metric-attention mechanism as candidates for the information propagation block. The differences among them lie in how they compute the pseudo-metric. Additionally, we replace the self-attention module in Transformer with metric-attention to test the compatibility of it with other commonly used modules. Experimental details can be found in Table \ref{Details of experiments} and Appendix \ref{Appendix_Experiments}.

\begin{table}[h]
    \centering
    \caption{Experimental details. $lr$: Learning rate. $\omega$: Weight decay. $d_{Q}$: Size of matrices $Q$ and $K$. $d_{mlp}$: Width of one-hidden-layer MLP in metric-attention.}
    \begin{tabular}{cccccccc}
    \toprule[0.8pt]
    Dataset & Model  & $lr$& $\omega$ & $d_{Q}$ & $d_{mlp}$\\
    \midrule[0.8pt]
    Moon & IPN & $10^{-3}$ & $10^{-4}$ & $10\times10$ &10\\
    Mnist & IPN & $10^{-4}$ & $10^{-4}$ & $100\times100$ & 100 \\
    Human segmentation & IPN &  $10^{-3}$ & $10^{-4}$ & $64\times64$ &64 \\
    Multi30K & Transformer  & $10^{-5}$ & $5\times 10^{-4}$ & $512\times512$& 512\\
    \bottomrule[0.8pt]

    \end{tabular}
    \vspace{1em} 
\label{Details of experiments}
\end{table}

Following previous studies, we measure the accuracy, robustness, and training efficiency of different attention mechanisms. For the classification and segmentation tasks, the accuracy is measured as the proportion of predicted labels that match the ground truth labels (or manually annotated labels). For the translation tasks, we use the Bleu metric to reflect their performance. Robustness is evaluated by the variance of accuracy at the end of training. Finally, we demonstrate the training efficiency of these mechanisms by plotting the training loss and test accuracy curves.

\begin{figure}[htbp]
   \center
   \includegraphics[scale=0.56]{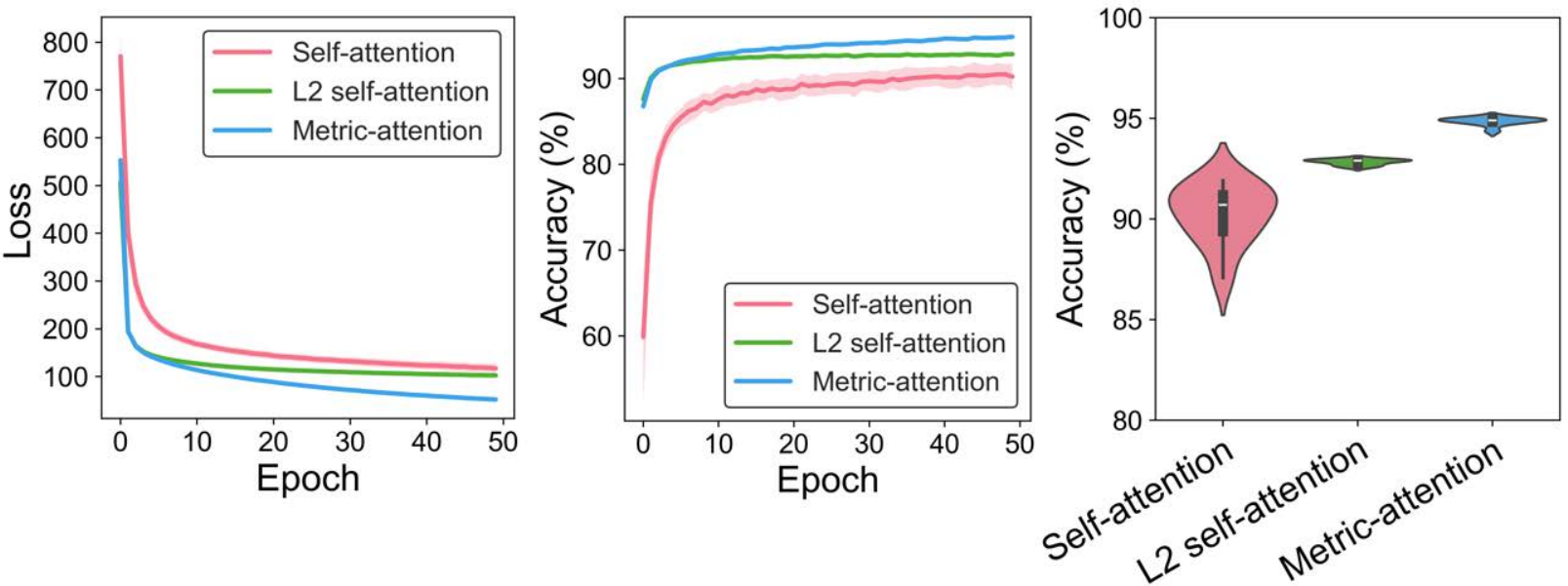}
   \caption{Experimental results on the MNIST dataset. The left subplot shows the loss curve during the training process. The middle subplot shows the testing accuracy during training. The right subplot is a violin plot of the test accuracy of three structures at the end of training.
}
   \label{fig:MNIST}
\end{figure}

\begin{figure}[h]
   \center
   \includegraphics[scale=0.80]{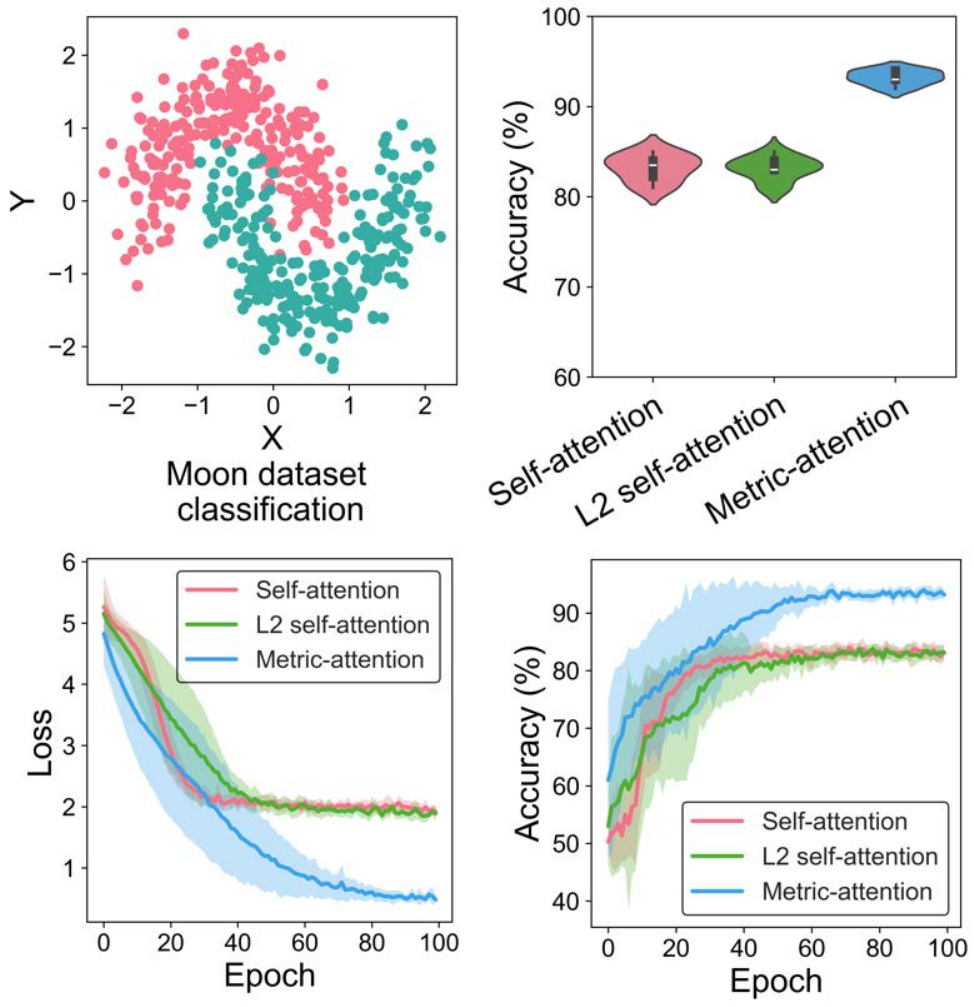}
   \caption{Experimental results on the Moon dataset. (top-left) Visualization of this dataset. (top-right) the testing accuracy of the three methods at the end of training. (bottom-left and bottom-right) illustration of the testing accuracy and loss of the three methods during the training process, respectively.
}
   \label{fig:Toy_dataset}
\end{figure}

\begin{figure}[h]
   \center
   \includegraphics[scale=0.60]{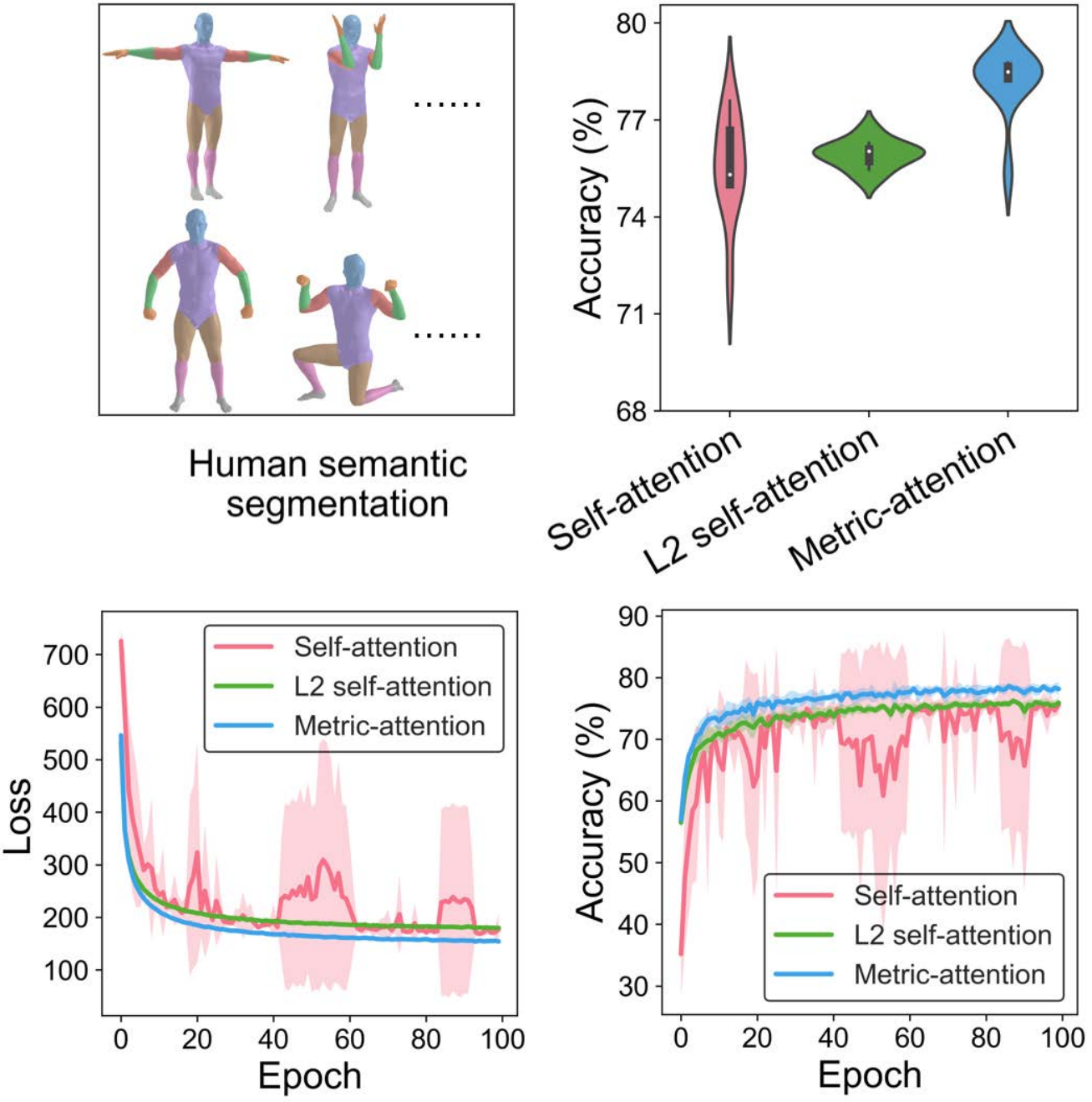}
   \caption{Evaluation of self-attention, L2 self-attention and metric-attention methods on the Human semantic segmentation dataset. (top-left) Visualization of four instances in the dataset. (top-right) the testing accuracy of the three methods at the end of training. (bottom-left and bottom-right) illustration of the testing accuracy and loss of the three methods during the training process, respectively.
}
   \label{fig:humanseg_dataset}
\end{figure}

\begin{figure}[h]
   \center
   \includegraphics[scale=0.60]{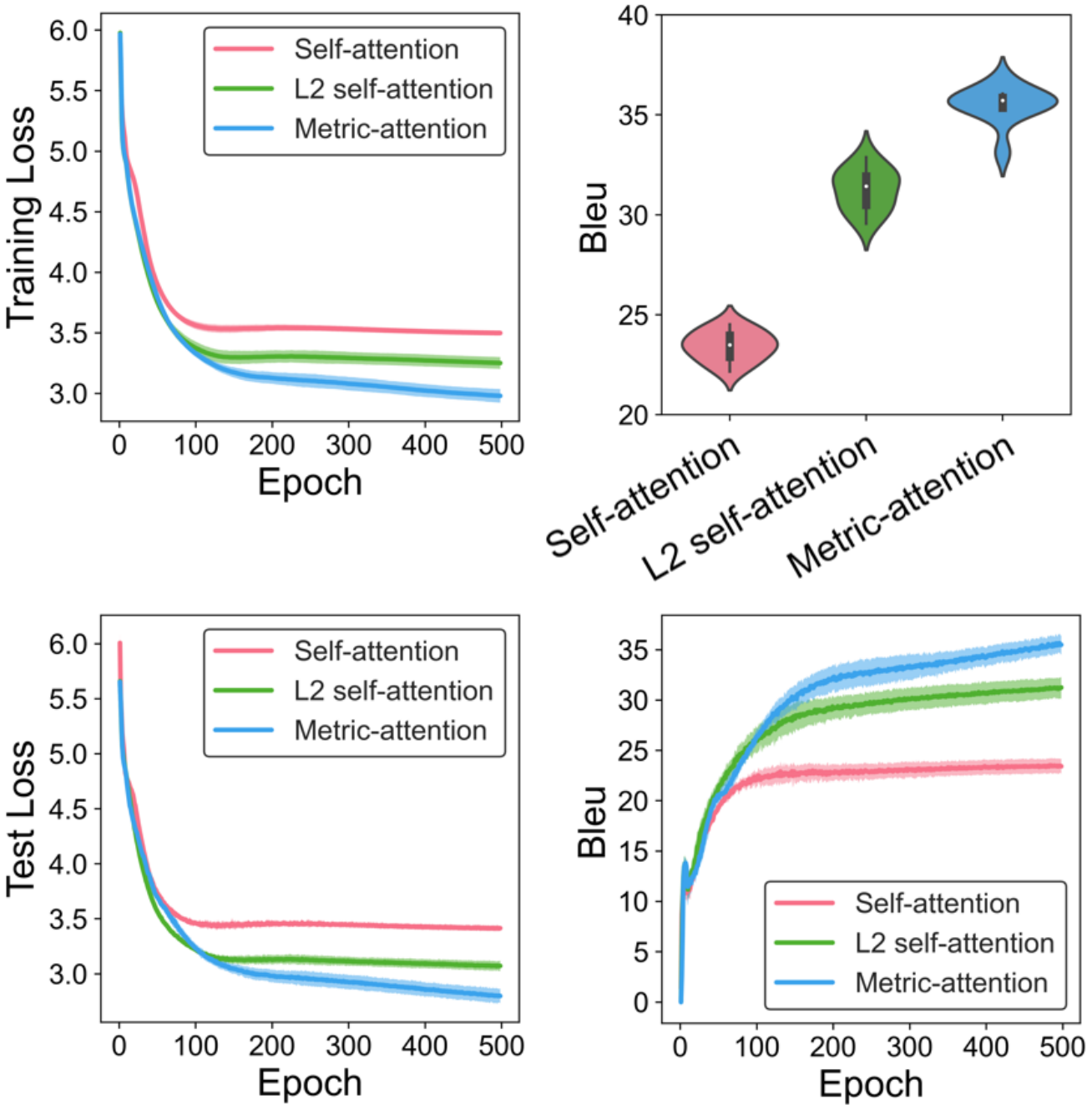}
   \caption{Experimental results on the Multi30k dataset. The top-left, bottom-left and bottom-right figure respectively illustrate the training loss, testing loss and Bleu scores during the training process. The top-right figure shows the Bleu scores of three information propagation methods at the end of the training.
}
   \label{fig:Multi30k}
\end{figure}

We conduct experiments on two vector-type data, i.e., the MNIST \citep{deng2012mnist} and Moon datasets to evaluate the performance of the self-attention, L2 self-attention, and metric-attention information propagation mechanisms (Figures \ref{fig:MNIST} and \ref{fig:Toy_dataset}). Additionally, we tested the metric-attention mechanism on a graph data (manifold data), i.e., the human semantic segmentation dataset \citep{maron2017convolutional,anguelov2005scape,bogo2014faust,giorgi2007shape,vlasic2008articulated} to assess its performance (Figure \ref{fig:humanseg_dataset}). In these experiments, the metric-attention mechanism resulted in lower classification loss during training and higher test accuracy at the end of training compared to the baselines, demonstrating its superior information processing capability and training efficiency. Moreover, violin plots of 10 repeated experiments illustrate the robustness of this mechanism.

We compared the performance of Transformers equipped with different information propagation modules on the Multi30k translation task. By examining training loss, testing loss (perplexity), and Bleu score curves, metric-attention demonstrates superior training efficiency, model performance and robustness (Figure \ref{fig:Multi30k}). This suggests that it is compatible with traditional architectures and demonstrates significant potential.

Compared to L2 self-attention and metric-attention, self-attention shows large variance in accuracy both during the training process and at the end of training (Figures \ref{fig:MNIST} and \ref{fig:humanseg_dataset}). Additionally, in Figure \ref{fig:humanseg_dataset}, the loss curve also exhibits significant variance and jitter. These observations suggest that the training process of self-attention may not be robust. \cite{kim2021lipschitz} indicate that the Lipschitz property of self-attention is poor, while the L2 form has a better Lipschitz property, which is beneficial for robust training. Furthermore, our analysis (Section \ref{general pseudo-metric setting}) shows that the information propagation mechanism of self-attention generally does not correspond to a continuous dynamical system; it only corresponds to a continuous dynamical system under rather ideal assumptions. In contrast, both L2 self-attention and metric-attention correspond to a continuous dynamical system under more relaxed conditions. These two reasons can explain the phenomenon of instability in the training process of the attention architecture observed in the experiments.

From a theoretical perspective, both L2 self-attention and metric-attention correspond to a continuous dynamical system determined by a learnable Riemannian metric. L2 self-attention obtains this Riemannian metric through linear transformation, while metric-attention uses a mini-MLP with residual connections to learn the Riemannian metric. This enables metric-attention to learn more complex metric relationships, thereby enhancing the performance of the algorithm. This explains the superior results achieved by the metric-attention architecture in various tasks (Figures \ref{fig:MNIST}, \ref{fig:Toy_dataset}, \ref{fig:humanseg_dataset} and \ref{fig:Multi30k}).

\section{Conclusion}
In this study, we first examine the techniques for similarity computation in classical machine learning algorithms such as manifold learning, clustering and supervised learning. We point out that attention mechanism is essentially a composition of an information propagation process based on a handcrafted similarity computation and a learnable pseudo-metric. This highlights the strong connection between attention mechanism and traditional algorithms. More importantly, we demonstrate the evolution of techniques over time: traditional algorithms rely on manual design for similarity computation, while neural networks represented by attention mechanism introduce learnable and adaptive techniques making them more flexible. Under the assumption that the pseudo-metric is a transformation of a metric function and some other assumptions, we utilize PDEs to explain the limit properties of attention mechanisms and translate them into heat equations for intuitive comprehension. For a general pseudo-metric, we fully account for its differences from a metric and provide a first-order analysis. This helps us intuitively understand the working principle of attention mechanism. That is, the features can be interpreted as updates using the nearest neighbors in terms of the pseudo-metric.
 
We conclude that the parameters of attention mechanism are designed to learn a pseudo-metric, which is used to compute pairwise similarities of data points. That implies we can consider training attention blocks as searching for a helpful pseudo-metric. It shares the same objective as metric learning. Thus, we integrate metric learning into attention mechanism. The difference between attention mechanism and metric learning lies in that metric learning often directly learns the metric through supervised approaches, while attention mechanism achieves this implicitly through propagation, making it more flexible. A significant advantage of attention mechanism is that it allows direct learning of a pseudo-metric through the labels provided by the task without requiring the metric to serve as the supervised signal. Moreover, different layers can learn different metrics, making pseudo-metric learning more powerful and flexible.

Previous researchers have attempted to understand deep neural networks (DNNs) using continuous dynamical systems. For example, some studies \citep{articleE,chen2018neural} have modeled residual neural networks \citep{he2016deep} as ordinary differential equations (ODEs). \cite{gai2021mathematical} utilized optimal transport to understand residual neural networks. \cite{song2020score} formalized denoising diffusion probabilistic models \citep{ho2020denoising} into stochastic differential equations (SDEs). Certain studies have linked graph neural networks with diffusion processes \citep{li2024generalized}. Some studies utilized measure theory to understand attention mechanism with residual connections \citep{geshkovski2023mathematical,sander2022sinkformers,vuckovic2020mathematical}. These studies share the commonality of interpreting the characteristic layer-by-layer updates of features in DNNs as the temporal dimension of information processing. Among these, some studies model the changes in features by dynamics of measures, while others model these as dynamics of functions. Here we adopt the latter perspective, distinguishing it from previous studies on attention mechanism \citep{geshkovski2023mathematical,sander2022sinkformers,vuckovic2020mathematical}. In addition, We reveal the connections between attention mechanism and classical algorithms that have not been addressed before.
 
However, this study has certain limitations. First, the limit properties require the satisfaction of three assumptions, which may not hold in engineering practice. Second, although attention mechanism is a crucial component of neural network architectures like Transformers, these networks are complex engineering products with many modules (e.g., multilayer perceptron, skip connection) that are not yet fully understood as a collaborative whole. Third, although we have tested the performance of the metric-attention mechanism, further experiments are needed to figure out how this mechanism performs across different domains and tasks.

Since attention mechanism can be viewed as a discretization of a heat equation with a learnable metric, a natural question arises. Can new network blocks be designed based on other PDEs that might be more universal or effective in specific domains? We anticipate that this could become a prominent area of research in the future.

\section*{Acknowledgments}
This work has been supported by the CAS Project for Young Scientists in Basic Research [No. YSBR-034], the National Key Research and Development Program of China [2019YFA0709501], and the National Natural Science Foundation of China [No. 12126605].
 
\bibliographystyle{unsrt}
\bibliography{attentionlimit}

\newpage
\appendix
\section*{Appendices}
\addcontentsline{toc}{section}{Appendices}
\renewcommand{\thesubsection}{\Alph{subsection}}

\section{Experiments and results}
\label{Appendix_Experiments}

\paragraph{Toy dataset}
We evaluated the performance of metric-attention by comparing it with self-attention and L2 self-attention using the Moon dataset. The Moon dataset is linearly inseparable. The training and testing data are generated using sklearn with a noise level of $0.2$ \citep{pedregosa2011scikit}. The training set size is $400$, and the test set size is $100$. We used the Adam optimizer and the cross-entropy loss function. For metric-attention, we employed the Tanh activation function.


\paragraph{MNIST handwritten digit database}
For the MNIST dataset, we applied the same settings as those used for the toy dataset, including the Adam optimizer and the cross-entropy loss function. For metric-attention, we employed a one-hidden-layer MLP with a width of $100$, the Tanh activation function, and a residual connection to learn a representation function as defined in Section \ref{Metric-attnetion}.
 
\paragraph{Human semantic segmentation}
The human semantic segmentation dataset \citep{maron2017convolutional} consists of numerous human 3D meshes along with their semantic segmentation. In each mesh, there are thousands of nodes and triangle faces and every face has a label according to their semantic meaning. We adopted a fixed linear aggregation layer to transform features of vertices into features of faces. We use the Adam optimizer and the cross-entropy loss function.

\paragraph{Multi30k dataset}
We used the Multi30k dataset \citep{elliott2016multi30k} to train Transformers with different information propagation blocks and evaluate their performance. We adopted the Adam optimizer, the ReduceLROnPlateau learning schedule, and the cross-entropy loss function. The number of heads is 8 and the number of layers is 6. 

\paragraph{Implementation} 
We implemented the deep neural networks using PyTorch \citep{paszke2019pytorch}. We conducted all experiments on a desktop computer with NVIDIA 2080Ti and 3.8 GHz AMD Ryzen 7 5800X 8-Core Process and 16 GB of memory and a computer with NVIDIA 3090Ti. We partly used the Diffusionnet \citep{sharp2022diffusionnet} to preprocess the human semantic segmentation dataset. We adapted the implementation for the Multi30k dataset from https://github.com/hyunwoongko/transformer.

\section{Proof of limit properties of metric setting}
\label{proof}
\paragraph{Proof of Theorem 2}
Suppose the data points $\{x_i\}_{i=1}^N$ are i.i.d. sampled from a distribution on a Riemannian manifold with the density function $p(x)$, then
\begin{align*}
    \frac{1}{\epsilon}\sum_{j=1}^n  L_{ij}f(x_j)\to \frac{1}{2}\left(\Delta f(x_i)+2\left\langle \frac{\nabla p}{p}(x_i), \nabla f(x_i)\right\rangle\right)
\end{align*}

\noindent\textbf{Proof:} (see \citet{coifman2006diffusion}). If Theorem 1 is acknowledged, Theorem 2 can be easily proven. 
Theorem 1 says:
\begin{align*}
    \frac{\int \exp{\left(-\frac{\lVert x-y\rVert^2}{2\epsilon}\right)}g(y)d\omega(y)}{\int \exp{\left(-\frac{\lVert x-y\rVert^2}{2\epsilon}\right)}d\omega(y)}=g(x)+\frac{\epsilon}{2}\Delta g(x)+{\rm Higher\,order}
\end{align*}
As a result, letting $g(y)=f(y)p(y)$ and $g(y)=p(y)$, we get the following estimate by taking a ratio:
\begin{equation}
\label{eqthm2}
\frac{\int \exp{\left(-\frac{\lVert x-y\rVert^2}{2\epsilon}\right)}f(y)p(y)d\omega(y)}{\int \exp{\left(-\frac{\lVert x-y\rVert^2}{2\epsilon}\right)}p(y)d\omega(y)}=\frac{f(x)p(x)+\frac{\epsilon}{2}\Delta \left(f(x)p(x)\right)+{\rm Higher\,order}}{p(x)+\frac{\epsilon}{2}\Delta p(x)+{\rm Higher\,order}}
\end{equation}
We know that
\begin{align*}
    \Delta\left(fp\right)=p\Delta f+f\Delta p+2\langle \nabla f,\nabla p\rangle
\end{align*}
Then Eq. (\ref{eqthm2}) can be written as:
\begin{align*}
    \frac{\int \exp{\left(-\frac{\lVert x-y\rVert^2}{2\epsilon}\right)}f(y)p(y)d\omega(y)}{\int \exp{\left(-\frac{\lVert x-y\rVert^2}{2\epsilon}\right)}p(y)d\omega(y)}&=f(x)+\frac{\frac{\epsilon}{2}\left(p(x)\Delta f+2\langle \nabla f,\nabla p\rangle \right)+{\rm Higher\,order}}{p(x)+\frac{\epsilon}{2}\Delta p(x)+{\rm Higher\,order}}\\
    &=f(x)+\frac{\epsilon}{2}\left(\Delta f(x)+2\left\langle \frac{\nabla p}{p}(x), \nabla f(x)\right\rangle\right)+{\rm Higher \,order}
\end{align*}
We complete the proof by a simple rearrangement and the law of large numbers.

$\hfill\square$
 
\paragraph{Proof of Theorem 3} If the aforementioned assumptions are satisfied, the dynamics of the feature of each data point in attention mechanism for information propagation is a first-order approximator (with respect to $\epsilon$) of a PDE:
\begin{align*}
    &\frac{dH}{dt}=\Delta_{g_\theta} H+2 \left\langle\frac{\nabla p}{p}, \nabla H\right\rangle\\
    &H(x,0)=H^{\rm old}(x)
\end{align*}
where $\Delta_{g_\theta}$ is the Laplacian-Beltrami operator of manifold $\mathcal{M}$ with the Riemannian metric given by neural network (Assumption 2) and $p$ is the density function (Assumption 3).

\noindent\textbf{Proof:} Denote the matrix of $\exp\{f_\theta(x_i,x_j)\}$ by $W$. Attention mechanism for information propagation can be described by:
\begin{align*}
   H^{\rm new}=D^{-1}W H^{\rm old}
\end{align*}
where $D$ is the degree matrix. Subtract $H^{\rm old}$ from both sides of the equation simultaneously, we get
\begin{align*}
   H^{\rm new}-H^{\rm old}=(D^{-1}W-I) H^{\rm old}
\end{align*}
By Theorem 2 and the analogous argument in \citep{coifman2006diffusion}, we know that $\big((D^{-1}W-I)H\big)(x_i) = \frac{\epsilon}{2}\cdot(\Delta H+ 2\frac{\nabla p}{p}\cdot \nabla H)(x_i)+{\rm Higher\,order}$. Thus, we have
\begin{align*}
   (H^{\rm new}-H^{\rm old})(x_i)=\frac{\epsilon}{2}\cdot\left(\Delta_{g_\theta} H^{\rm old}+ \frac{\nabla p}{p}\cdot \nabla H^{\rm old}\right)(x_i)+{\rm Higher\,order}
\end{align*}
which is the formula of the Euler method of PDE ($n\to\infty$):
\begin{align*}
    \frac{dH}{dt}=\Delta_{g_\theta} H+ \frac{\nabla p}{p}\cdot \nabla H
\end{align*}
$\hfill\square$
 
\paragraph{Proof of Theorem 4}
If the dimension $n\neq2$, then there exists a metric $\tilde{g}$ such that the At operator equals a Laplacian-like operator, i.e.,
\begin{align*}
     \Delta_g+2\left\langle\frac{\nabla p}{p}, \nabla \cdot \right\rangle=f\Delta_{\tilde{g}}
\end{align*}
where $f=p^{4/n-2}$.
 
\noindent\textbf{Proof:} Given a Riemannian metric $g=g_{ij}$, the Laplacian-Beltrami operator can be calculated by the following formula:
\begin{align*}
\Delta_g=\frac{1}{\sqrt{|g|}}\partial_i(\sqrt{|g|}g^{ij}\partial_j)    
\end{align*}
Given another Riemannian metric $\tilde{g}=e^{2\lambda}g=e^{2\lambda}g_{ij}$, which is a conformal metric of $g$, we directly calculate its Laplacian:
\begin{align*}
\Delta_{\tilde{g}}&=\frac{1}{\sqrt{|\tilde{g}|}}\partial_i(\sqrt{|\tilde{g}|}\tilde{g}^{ij}\partial_j)\\
&=\frac{1}{e^{n\lambda}\sqrt{|g|}}\partial_i(e^{(n-2)\lambda}\sqrt{|g|}g^{ij}\partial_j)\\
&=e^{-2\lambda}\frac{1}{\sqrt{|g|}}\partial_i(\sqrt{|g|}g^{ij}\partial_j)+(n-2)e^{-2\lambda}(\partial_i\lambda) g^{ij}\partial_j\\
&=e^{-2\lambda}\left(\Delta_g+(n-2)\langle \nabla \lambda,\nabla\cdot\rangle\right)
\end{align*}
We have
\begin{align*}
   e^{2\lambda}\Delta_{\tilde{g}}=\Delta_g+\langle \nabla (n-2)\lambda,\nabla\cdot\rangle
\end{align*}
As $n\ge3$, $n-2\neq0$. Let $\lambda=\frac{2\log p}{n-2}$, we have:
\begin{align*}
   p^{4/n-2}\Delta_{\tilde{g}}=\Delta_g+2\left\langle  \frac{\nabla p}{p},\nabla\cdot\right\rangle
\end{align*}
$\hfill\square$
 
\paragraph{Dynamic of heat diffusion} Heat diffusion on the manifold $(\mathcal{M},\tilde{g})$ with the specific heat capacity $c$, thermal conductivity $k$ and material density $\rho$ has dynamic:
\begin{align*}
   \frac{du}{dt}=\frac{1}{c\rho}\nabla \cdot(k\nabla u)
\end{align*}
When $k=1,\rho=1$ and $c=f^{-1}$, this dynamic is the same as the dynamic of attention.
 
\noindent\textbf{Proof:} By the Fourier's law, the flow of heat can be described by a vector field:
\begin{align*}
   q=-k\nabla u
\end{align*}
Denote the heat energy at point $x$ and time $t$ by $Q(x,t)$, we know that the change of temperature is proportional to the change of heat energy. To be specific:
\begin{align*}
   \frac{\partial Q}{\partial t}=c\rho\frac{\partial u}{\partial t}
\end{align*}
Additionally, we know that the change in heat energy is equal to the net heat flux:
\begin{align*}
   \frac{\partial Q}{\partial t}= -{\rm div}q
\end{align*}
Therefore, we have:
\begin{align*}
   \frac{\partial u}{\partial t}=\frac{1}{c\rho}\nabla \cdot(k\nabla u)
\end{align*}
Let $k=1,\rho=1$ and $c=f^{-1}$, we have
\begin{align*}
   \frac{\partial u}{\partial t}=f\Delta u
\end{align*}
$\hfill\square$

\section{Proof of limit properties in general pseudo-metric setting}
\label{proof_of_general pseudo-metric_setting}

\paragraph{Example 3} We shall solve the following problem:
\begin{align*}
    \text{minimize:} & \quad \sum a_ix_i'y_i' \\
\text{subject to:} & \quad \sum y'^{2}_i = 1
\end{align*}
The derivatives of Laplacian $L(y',\lambda)=\sum a_ix_i'y_i'-\lambda(\sum y'^{2}_i-1)$ are:
\begin{align*}
    &\frac{\partial L}{\partial y'_i}=a_ix'_i-2\lambda y'_i\\
    &\frac{\partial L}{\partial \lambda}=\sum y'^{2}_i-1
\end{align*}
By letting these derivatives equal $0$, we have:
\begin{align*}
    &\lambda=\frac{\sqrt{\sum a_i^2x'^{2}_i}}{2}\\
    &y'_i=\frac{a_ix'_i}{\sqrt{\sum a_i^2x'^{2}_i}}
\end{align*}
Since for an ellipsoid, there exists a linear transform to transform it into a unit sphere, we complete the proof.

$\hfill\square$
\paragraph{The first-order analysis for limit properties in general pseudo-metric setting:}
We firstly introduce our assumptions:
\paragraph{Formalization}
We focus on the information propagation process of attention mechanism:
\begin{align*}
   H^{\rm new}=SH^{\rm old}
\end{align*}
where
$S_{ij}=\frac{\exp \big(-f_\theta(x_i,x_j)\big)}{\sum_k \exp\big(-f_\theta(x_i,x_k)\big)}$. 
We reformulate the similarity matrix of attention mechanism $S$ as $S_\epsilon$:
\begin{align*}
    S_{\epsilon,ij}=\frac{\exp \big(-\frac{f_\theta(H^{\rm old}_i,H^{\rm old}_j)}{2\epsilon}\big)}{\sum_k \exp\big(-\frac{f_\theta(H^{\rm old}_i,H^{\rm old}_k)}{2\epsilon}\big)}
\end{align*}
where $\epsilon/2$ represents the time scale. In attention mechanism, the updated representation is given by $H^{\rm new}=S_{\epsilon}H^{\rm old}$ where $ \epsilon=\frac{1}{2}$. Based on this, we suppose:
\paragraph{Assumption 1:} $\epsilon$ is sufficiently small.
 
\paragraph{Assumption 2:}
We adhere to the manifold hypothesis, which posits that: data lie on a compact, connected Riemannian manifold $\mathcal{M}$ and they are i.i.d. sampled from a random variable $X$ whose density function $p(x)$ with respect to the volume element $dx$. Suppose we observe data in the Euclidean space by an embedding $H^{\rm old}(x)$. Besides, we suppose that we have sufficiently many data.

\paragraph{Assumption 3: Regularity conditions}
We assume that $f_\theta$ is a smooth function and $A_x={\rm argmin}_y f_\theta(x,y)$ is a compact geodesic (flat) submanifold of $\mathcal{M}$ for all $x\in \mathcal{M}$ for simplicity. In addition, we assume:
\begin{align*}
    \dim\left(\ker\circ \nabla_y^2f_\theta(x,y)\right)=\dim A_x,\forall y\in A_x 
\end{align*}
i.e., the Hessian matrix of $f_\theta$ is non-degenerate in the normal direction of the manifold $A_x$.

Denote ${\rm argmin} f_\theta(x,\cdot)$ by $A_x$ and denote its $\delta$ neighbors $\{y+z\in\mathcal{M},y\in A_x,z\in B(0,\delta)\}$ by $A_{x,\delta}$. Denote $\min f_\theta(x,\cdot)$ as $\rho(x)$.
Since $A_{x}$ is a geodesic submanifold, we can use $\exp$ to get the chart of tubular neighborhood $\{\psi,U\times V\}$: 
\begin{align*}
    \psi:U\times V\to \mathbb{R}^{n+m}
\end{align*}
where $n$ and $m+n$ are the dimensionalities of $A_{x}$ and $\mathcal{M}$, respectively. $\forall (u,0)\in U\times V, \psi(u,0)\in A_{x},\psi(U,V)\subset A_{x,\delta}$ and the distance from $\psi(u,v)$ to $A_x$ equals $\lVert v \rVert$.

We use $y$ to represent a point on $\mathcal{M}$ and $y'$ to represent a point on $A_x$. We denote $f\circ \psi$ by $\tilde{f}$, $H\circ \psi$ by $\tilde{H}$, and $p\circ \psi$ by $\tilde{p}$. 

Since the Hessian of $f$ with respect to to $v$ is non-degenerate, there exists constants $C_1,C_2$ and $\delta_1$ sufficiently small, s.t., 
\begin{align*}
    \rho(x)+ C_1\lVert v\rVert^2\le \tilde{f}(u,v) \le \rho(x)+ C_2\lVert v\rVert^2,\, \forall \lVert v \rVert^2\le\delta_1
\end{align*}

Besides, since $\exp{\left(\frac{-\lVert v\rVert^2}{2\epsilon}\right)}$ decreases exponentially, we have:
\begin{align*}
    \int \exp\left(\frac{-\Vert v\Vert^2}{2\epsilon}\right)dv\simeq \int_{B_{C\sqrt{\epsilon}}}\exp\left(\frac{-\Vert v\Vert^2}{2\epsilon}\right)dv
\end{align*}
where $B_{C\sqrt{\epsilon}}$ is a ball with radius $C\sqrt{\epsilon}$. Therefore, let $U=B_{\delta},\,\delta=C\epsilon^{1/2}$, we have:
\begin{align*}
    \exp{\left(\frac{-\rho(x)}{2\epsilon}\right)}\int \exp\left(\frac{-\tilde{f}_\theta(u,v)}{2\epsilon}\right)dv\simeq \exp{\left(\frac{-\rho(x)}{2\epsilon}\right)} \int_{B_{C\sqrt{\epsilon}}}\exp\left(\frac{- \tilde{f}_\theta(u,v)}{2\epsilon}\right)dv
\end{align*}

\begin{Lemma}
\label{lemma_volum}
By $\psi$, we parameterize $A_{x,\delta}$ and have:
\begin{align*}
    dy=\left(1-\frac{1}{6}\sum R_{kl}(y')v_kv_l\right)dy'dv+\mathcal{O}\left(\epsilon^{3/2}\right)
\end{align*}
where $R_{ij}$ is the Ricci curvature of $\mathcal{M}$ at $y$, $dy$ is the volume element of $\mathcal{M}$, $dy'$ is the volume element of $A_x$, $y'=\psi(u,0)$.
\end{Lemma}

\noindent\textbf{Proof:} In normal coordinates, the Taylor expansion of $g_{ij}$ is:
\begin{align*}
    g_{ij}=\delta_{ij}-\frac{1}{3}\sum_{k,l} R_{ikjl}v_kv_l+\mathcal{O}(\lVert v\rVert^3)
\end{align*}
where $R_{ikjl}$ is the sectional curvature. As a result,
\begin{align*}
    \sqrt{\det g}=1-\sum_{k,l}\frac{1}{6}R_{kl}v_kv_l+\mathcal{O}(\lVert v\rVert^3)
\end{align*}
By this calculation, we complete the proof.

$\hfill\square$

Expand $\tilde{f}$ up to the fourth order: by the Taylor series, for any $y\in\psi(U)$,
\begin{align*}
    \tilde{f}_\theta(u,v)-\tilde{f}_\theta(u,0) =  \sum_{k,l} \frac{c_{kl}}{2}v_kv_l+\sum_{k,l,m}\frac{d_{klm}}{3!}v_kv_lv_m+\sum_{k,l,m,n}\frac{e_{klmn}}{4!}v_kv_lv_mv_n+\mathcal{O}\left(\epsilon^{5/2}\right)
\end{align*}
where $c_{kl}=\frac{\partial^2 \tilde{f}_\theta}{\partial v_k\partial v_l},d_{klm}=\frac{\partial^3 \tilde{f}_\theta}{\partial v_k\partial v_l\partial v_m},e_{klmn}=\frac{\partial^4 \tilde{f}_\theta}{\partial v_k\partial v_l\partial v_m \partial v_n}$, where $c,d,e$ are functions of $u$ (or functions of $y\in A_{x}$).

\textbf{Note 1}: For smooth functions, we can find $C$ such that $\mathcal{O}(\epsilon^{5/2})$ is bounded by $C\epsilon^{5/2}$. This is because of the compactness of $A_x$, there exists a $B_{\epsilon_1}\times B_{\epsilon_2}$ such that $\forall v\in B_{\epsilon_1}$:
\begin{align*}
     \Big|\tilde{f}_\theta(u,v)-\tilde{f}_\theta((u,0) - \sum_{k,l} \frac{c_{kl}}{2}v_kv_l-\sum_{k,l,m}\frac{d_{klm}}{3!}v_kv_lv_m-&\sum_{k,l,m,n}\frac{e_{klmn}}{4!}v_kv_lv_mv_n \Big| \le C\Vert v\Vert^5
\end{align*}
By the compactness, there exists a $C$ independent of $u$ (or $y'$).
\begin{Lemma}
Denote $\left(1-\frac{1}{6}\sum R_{kl}(y')v_kv_l\right)dv$ by $dy^{\perp}$. 
\label{taylor}
For $y\in A_{x,\delta}$, $y=\psi(u,0)$,
\begin{align*}
    &\int_{u\times V}\exp{\left(\frac{-\tilde{f}_\theta(u,v)}{2\epsilon}\right)}\tilde{g}(u,v)dy^{\perp}\\
    = &\epsilon^{m/2}\exp\left(\frac{\rho(x)}{2\epsilon}\right)\left(m_0g(y)+\epsilon\langle V_{f_\theta},\nabla_v g\rangle+\epsilon E(y)g(y)+\epsilon\Delta_{c^{-1},v}g+\mathcal{O}\left(\epsilon^{3/2}\right)\right)
\end{align*}
where
\begin{align*}
    &m_0=\int \exp\left(\frac{-\sum \frac{c_{kl}}{2}v_kv_l}{2}\right)dv,\quad \Delta_{c^{-1},v}g=\sum_{k}c^{-1}_{kl}\frac{\partial^2 g}{\partial v_kv_l}\\
    &V_{f_\theta}=\sum_i\int\exp\left(\frac{-\sum \frac{c_{kl}}{2}v_kv_l}{2}\right) \frac{-\sum_{k,l,m}\frac{d_{klm}}{3!}v_kv_lv_mv_i}{2}dv\,\frac{\partial}{\partial v_i}\\
    &E(y)=\int \exp\left(\frac{-\sum \frac{c_{kl}}{2}v_kv_l}{2}\right)\frac{\sum_{k,l,m,n,o,p}\frac{d_{klm}d_{nop}}{3!\times 3!}v_kv_lv_mv_nv_ov_p}{8}dv\\
    &+\int\exp\left(\frac{-\sum \frac{c_{kl}}{2}v_kv_l}{2}\right)\frac{-\sum R_{kl} v_kv_l}{3} dv\\
    &+\int \exp\left(\frac{-\sum \frac{c_{kl}}{2}v_kv_l}{2}\right)\frac{-\sum_{k,l,m,n}\frac{e_{klmn}}{4!}v_kv_lv_mv_n}{2} dv
\end{align*}
\end{Lemma}

\noindent\textbf{Proof:} By the definition, we have:
\begin{align*}
    &\int_{u\times V}\exp{\left(\frac{-\tilde{f}_\theta(u,v)}{2\epsilon}\right)}\tilde{g}(u,v)dy^{\perp}\\
    =\exp\left(\frac{\rho(x)}{2\epsilon}\right)&\int\exp\left(\frac{-\sum_{k,l} \frac{c_{kl}}{2!}v_kv_l-\sum_{k,l,m}\frac{d_{klm}}{3!}v_kv_lv_m-\sum_{k,l,m,n}\frac{e_{klmn}}{4!}v_kv_lv_mv_n}{2\epsilon}\right)\\
    &\times\left(g(y)+\sum v_k\frac{\partial \tilde{g}}{\partial v_k}+\frac{1}{2}\sum v_k v_l\frac{\partial^2 \tilde{g}}{\partial v_k\partial v_l}\right) \times \left(1-\frac{1}{6}\sum R_{kl}v_kv_l\right)dv+\mathcal{O}\left(\epsilon^{3/2}\right)
\end{align*}
By the Taylor series, we have:
\begin{align*}
    &\exp\left(\frac{-\sum_{k,l} \frac{c_{kl}}{2!}v_kv_l-\sum_{k,l,m}\frac{d_{klm}}{3!}v_kv_lv_m-\sum_{k,l,m,n}\frac{e_{klmn}}{4!}v_kv_lv_mv_n}{2\epsilon}\right)\\
    =&\exp\left(\frac{-\sum_{k,l} \frac{c_{kl}}{2}v_kv_l}{2\epsilon}\right)+\exp\left(\frac{-\sum_{k,l} \frac{c_{kl}}{2}v_kv_l}{2\epsilon}\right)\\
    &\times\left(\frac{-\sum_{k,l,m}\frac{d_{klm}}{3!}v_kv_lv_m}{2\epsilon}+\frac{-\sum_{k,l,m,n}\frac{e_{klmn}}{4!}v_kv_lv_mv_n}{2\epsilon}\right)\\
    &+\exp\left(\frac{-\sum_{k,l} \frac{c_{kl}}{2!}v_kv_l}{2\epsilon}\right)\frac{\sum_{k,l,m,n,o,p}\frac{d_{klm}d_{nop}}{3!\times 3!}v_kv_lv_mv_nv_ov_p}{2\times 4\epsilon^2}
    +\mathcal{O}\left(\epsilon^{3/2}\right)
\end{align*}
Since the integral of an odd function equals zero, we have:
\begin{align*}
    &\int_{u\times V}\exp{\left(\frac{-\tilde{f}_\theta(u,v)}{2\epsilon}\right)}\tilde{g}(u,v)dy^{\perp}\\
    =&\exp\left(\frac{\rho(x)}{2\epsilon}\right)\int \exp\left(\frac{-\sum \frac{c_{kl}}{2}v_kv_l}{2\epsilon}\right)g(y)dv\\
    &+\sum_i\int \exp\left(\frac{-\sum \frac{c_{kl}}{2}v_kv_l}{2\epsilon}\right)\frac{-\sum_{k,l,m}\frac{d_{klm}}{3!}v_kv_lv_mv_i}{2\epsilon}\frac{\partial \tilde{g}}{\partial v_i}dv\\
    &+\int \exp\left(\frac{-\sum \frac{c_{kl}}{2}v_kv_l}{2\epsilon}\right)\frac{-\sum_{k,l,m,n}\frac{e_{klmn}}{4!}v_kv_lv_mv_n}{2\epsilon} g(y)dv\\
    &+\int\exp\left(\frac{-\sum \frac{c_{kl}}{2}v_kv_l}{2\epsilon}\right)\frac{1}{2}\sum v_kv_l\frac{\partial^2 \tilde{g}}{\partial v_kv_l}dv\\
    &+\int\exp\left(\frac{-\sum \frac{c_{kl}}{2}v_kv_l}{2\epsilon}\right)\frac{-1}{6}\sum R_{kl} v_kv_l g(y)dv\\
    &+\int \exp\left(\frac{-\sum \frac{c_{kl}}{2}v_kv_l}{2\epsilon}\right)\frac{\sum_{k,l,m,n,o,p}\frac{d_{klm}d_{nop}}{3!\times 3!}v_kv_lv_mv_nv_ov_p}{8\epsilon^2}g(y)dv
    +\mathcal{O}\left(\epsilon^{3/2}\right)
\end{align*}
Finally, we have:
\begin{align*}
    &\int_{u\times V}\exp{\left(\frac{-\tilde{f}_\theta(u,v)}{2\epsilon}\right)}\tilde{g}(u,v)dy^{\perp}\\
    = &\epsilon^{m/2}\exp\left(\frac{\rho(x)}{2\epsilon}\right)\left(m_0g(y)+\epsilon\langle V_{f_\theta},\nabla_v g\rangle+\epsilon E(y)g(y)+\epsilon\Delta_{c^{-1},v}g+\mathcal{O}\left(\epsilon^{3/2}\right)\right)
\end{align*}
$\hfill\square$

The following proposition gives the first-order expansion of information propagation of attention mechanism when $A_x=\{y'\}$, which could be regarded as a generalization of the heat kernel approximation.
\begin{Proposition}
\begin{align*}
    &\frac{\int_{u\times V}\exp{\left(\frac{-\tilde{f}_\theta(u,v)}{2\epsilon}\right)}\tilde{g}\tilde{p}(u,v)dy^{\perp}}{\int_{u\times V}\exp{\left(\frac{-\tilde{f}_\theta(u,v)}{2\epsilon}\right)}\tilde{p}(u,v)dy^{\perp}}=g(y')+\frac{\epsilon}{m_0}\left(\left\langle V_{f_\theta}+2\nabla_v\log p\cdot c^{-1},\nabla_v g\right\rangle+\Delta_{c^{-1},v} g\right)(y')+\mathcal{O}\left(\epsilon^{3/2}\right)
\end{align*}
\end{Proposition}

\noindent\textbf{Proof:} By Lemma \ref{taylor}, we have:
\begin{align*}
    &\epsilon^{-m/2}\int_{u\times V}\exp{\left(\frac{-\tilde{f}_\theta(u,v)}{2\epsilon}\right)}\tilde{g}\tilde{p}(u,v)dy^{\perp}\\
    = &\exp\left(\frac{\rho(x)}{2\epsilon}\right)\left(m_0gp(y')+\epsilon\langle V_{f_\theta},\nabla_v gp\rangle(y')+\epsilon Epg(y')+\epsilon\Delta_{c^{-1},v}gp(y')+\mathcal{O}\left(\epsilon^{3/2}\right)\right)\\
    =&\exp\left(\frac{\rho(x)}{2\epsilon}\right)\Big(m_0gp(y')+\epsilon\langle V_{f_\theta},\nabla_v g\rangle p(y')+\langle V_{f_\theta},\nabla_v p\rangle g(y')+\epsilon Epg(y')\\
    +&\epsilon p\Delta_{c^{-1},v}g(y')+\epsilon g\Delta_{c^{-1},v}p(y')+2\langle\nabla_v pc^{-1},\nabla_v g \rangle(y')+\mathcal{O}\left(\epsilon^{3/2}\right)\Big)
\end{align*}
Besides, we have:
\begin{align*}
    &\epsilon^{-m/2}\int_{u\times V}\exp{\left(\frac{-\tilde{f}_\theta(u,v)}{2\epsilon}\right)}\tilde{p}(u,v)dy^{\perp}\\
    =&\exp\left(\frac{\rho(x)}{2\epsilon}\right)\left(m_0p(y')+\epsilon\langle V_{f_\theta},\nabla_v p\rangle(y')+\epsilon Ep(y')+\epsilon\Delta_{c^{-1},v}p(y')+\mathcal{O}\left(\epsilon^{3/2}\right)\right)
\end{align*}
We complete the proof by taking a ratio.

$\hfill\square$

Denote $\frac{1}{m_0}\left(\left\langle V_{f_\theta}+2\nabla\log p\cdot c^{-1},\nabla g\right\rangle+\Delta_{c^{-1}} g\right)$ by ${\rm At}_{f_\theta,p}g$.
By Proposition 1 and the law of large numbers, if $A_x=\{y'\}$, we have:
\begin{align*}
    H^{\rm new}(x) = H^{\rm old}(y')+\epsilon {\rm At}_{f_\theta,p}H^{\rm old}(y') + {\rm Higher\,order}
\end{align*}
which complete the proof of Theorem \ref{general_thm_y'}. Generally, we can depict the information propagation of attention mechanism by the following theorem:
\begin{Theorem}[The first-order expansion for general pseudo-metric]
As $\epsilon\to 0$, the information propagation of attention mechanism has the first-order expansion:
\begin{align*}
    & H^{\rm new}(x) =\mathbb{E}^{Y\sim pm_0}(H^{\rm old}(Y)|Y\in A_x)\\
    &+\epsilon\left( +\mathbb{E}^{Y\sim pm_0}\left(\frac{h_0}{pm_0}(Y)|Y\in A_x\right)-\mathbb{E}^{Y\sim pm_0}\left(\frac{h_1}{pm_0}(Y)|Y\in A_x\right)\mathbb{E}^{Y\sim m_0p}(H^{\rm old}(Y)|Y\in A_x)\right)\\
    &+ {\rm Higher\,order}
\end{align*}
where
\begin{align*}
    h_0(y')&=EpH^{\rm old}(y')+\left\langle V_{f_\theta}, \nabla_v p\right\rangle H^{\rm old}(y')+\left\langle V_{f_\theta},\nabla_v H^{\rm old} \right\rangle p(y')\\
    &+p\Delta_{c^{-1},v}H^{\rm old}(y')+2\left\langle\nabla_v H^{\rm old}c^{-1},\nabla p \right\rangle(y')+H^{\rm old}\Delta_{c^{-1},v}p(y')\\
    h_1(y')&=\langle V_{f_\theta},\nabla p\rangle(y')+\Delta_{c^{-1},v}p(y')+Ep(y')
\end{align*}
\end{Theorem}
Here, $\nabla_v$ means to take derivative in the direction of normal space of $A_x$. $\Delta_{c^{-1},v}$ means to take weighted sum of the second-order derivatives by $c^{-1}$ in the direction of normal space of $A_x$ (Lemma \ref{taylor}).

\noindent\textbf{Proof:}
\begin{align*}
    &\int_\mathcal{M}\exp{\left(\frac{-f_\theta(x,y)}{2\epsilon}\right)}H^{\rm old}(y)p(y)dy\\
    =&\int_{A_{x,\delta}}\exp{\left(\frac{-f_\theta(x,y)}{2\epsilon}\right)}H^{\rm old}(y)p(y)dy + {\rm Higher\,order}\\
    =&\int_{A_x}\int_{V}\exp{\left(\frac{-\tilde{f}_\theta(u,v)}{2\epsilon}\right)}\tilde{H}^{\rm old}(u,v)\tilde{p}(u,v)\,dy^{\perp}dy' + {\rm Higher\,order}\\
    =&\int_{A_x}\int_{V}\exp{\left(\frac{-\tilde{f}_\theta(u,v)}{2\epsilon}\right)}\tilde{H}^{\rm old}(u,v)\left(\tilde{p}+\sum v_k\frac{\partial \tilde{p}}{\partial v_k}+\sum v_kv_l\frac{\partial^2 \tilde{p}}{\partial v_k\partial v_l}\right)(u,0)\,dy^{\perp}dy' + {\rm Higher\,order}
\end{align*}
Let $H^{\rm old}=1$, we have:
\begin{align*}
    &\int_\mathcal{M}\exp{\left(\frac{-f_\theta(x,y)}{2\epsilon}\right)}p(y)dy\\
    =&\int_{A_x}\int_{V}\exp{\left(\frac{-\tilde{f}_\theta(u,v)}{2\epsilon}\right)}\left(\tilde{p}+\sum v_k\frac{\partial \tilde{p}}{\partial v_k}+\sum v_kv_l\frac{\partial^2 \tilde{p}}{\partial v_k\partial v_l}\right)(u,0)\,dy^{\perp}dy' + {\rm Higher\,order}
\end{align*}
By Lemma \ref{taylor}, we have:

\begin{align*}
    \frac{\int_{A_x}\int_{V}\exp{\left(\frac{-\tilde{f}_\theta(u,v)}{2\epsilon}\right)}\tilde{H}^{\rm old}(u,v)\tilde{p}(u,v) \,dy^{\perp}dy'}{\int_{A_x}\int_{V}\exp{\left(\frac{-\tilde{f}_\theta(u,v)}{2\epsilon}\right)}\tilde{p}(u,v) \,dy^{\perp}dy'}=\frac{\int_{A_x}  pm_0H^{\rm old}(y')dy'+\epsilon\int_{A_x} h_0(y')dy'+{\rm Higher\,order}}{\int_{A_x} pm_0(y') dy' + \epsilon \int_{A_x} h_1(y') dy' +{\rm Higher\,order}}
\end{align*}
where
\begin{align*}
    h_0(y')&=EpH^{\rm old}(y')+\left\langle V_{f_\theta}, \nabla_v p\right\rangle H^{\rm old}(y')+\left\langle V_{f_\theta},\nabla_v H^{\rm old} \right\rangle p(y')\\
    &+p\Delta_{c^{-1},v}H^{\rm old}(y')+2\left\langle\nabla_v H^{\rm old}c^{-1},\nabla p \right\rangle(y')+H^{\rm old}\Delta_{c^{-1},v}p(y')\\
    h_1(y')&=\langle V_{f_\theta},\nabla p\rangle(y')+\Delta_{c^{-1},v}p(y')+Ep(y')
\end{align*}
As a result,
\begin{align*}
    &\frac{\int_{A_x}\int_{V}\exp{\left(\frac{-\tilde{f}_\theta(u,v)}{2\epsilon}\right)}\tilde{H}^{\rm old}(u,v)\tilde{p}(u,v) \,dy'dy^{\perp}}{\int_{A_x}\int_{V}\exp{\left(\frac{-\tilde{f}_\theta(u,v)}{2\epsilon}\right)}\tilde{p}(u,v) \,dy'dy^{\perp}}\\
    =& \frac{\int pm_0H^{\rm old}(y')\,dy'}{\int pm_0(y')\,dy'}+\frac{\epsilon\int h_0(y') dy'-\epsilon \int h_1(y')dy'\frac{\int pm_0H^{\rm old}(y')dy'}{\int pm_0(y')dy'}}{\int pm_0(y')dy'}+{\rm Higher\,order}
\end{align*}
where
\begin{align*}
    &\frac{\int_{A_x} pm_0H^{\rm old}(y')\,dy'}{\int_{A_x} pm_0(y')\,dy'}=\mathbb{E}^{Y\sim pm_0}(H^{\rm old}(Y)|Y\in A_x)\\
    &\frac{\int_{A_x} h_0(y')\,dy'}{\int_{A_x} pm_0(y')\,dy'}=\mathbb{E}^{Y\sim pm_0}\left(\frac{h_0}{pm_0}(Y)|Y\in A_x\right)\\
    &\frac{\int_{A_x} h_1\,dy'}{\int_{A_x} pm_0(y')\,dy'}=\mathbb{E}^{Y\sim pm_0}\left(\frac{h_1}{pm_0}(Y)|Y\in A_x\right)
\end{align*}
We complete the proof by the law of large numbers.

$\hfill\square$

\end{document}